\documentclass[letterpaper,twocolumn,fleqn]{article}

\usepackage{ist}
\usepackage[latin9]{inputenc}
\usepackage{graphicx,subfig,epsfig}
\usepackage{color}
\usepackage{balance}
\usepackage{float}
\usepackage{amsmath, amsfonts, amssymb, amsthm}
\usepackage{epstopdf}

\newcommand{\x}     {{\bf x}}
\newcommand{\y}     {{\bf y}}
\newcommand{\wx}    {\widehat{\bf x}}
\newcommand{\z}     {{\bf z}}

\newcommand{\RR}    {\mathbb{R}}
\newcommand{\hr}    {\hat{r}}

\title{Autoencoder with recurrent neural networks for video forgery detection}

\author{Dario D'Avino, Davide Cozzolino, Giovanni Poggi, Luisa Verdoliva \\
	DIETI, University Federico II of Naples, 80125 Naples, Italy}

\date{} 

\begin{document}

\maketitle

\thispagestyle{empty} 


\begin{abstract}
	Video forgery detection is becoming an important issue in recent years,
	because modern editing software provide powerful and easy-to-use tools to manipulate videos.
	In this paper we propose to perform detection by means of deep learning, with an architecture based on autoencoders and recurrent neural networks.
	A training phase on a few pristine frames allows the autoencoder to learn an intrinsic model of the source.
	Then, forged material is singled out as anomalous, as it does not fit the learned model, and is encoded with a large reconstruction error.
	Recursive networks, implemented with the long short-term memory model, are used to exploit temporal dependencies.
	Preliminary results on forged videos show the potential of this approach.
\end{abstract}

\section{Introduction}
\label{sec:intro}
An ever increasing share of the traffic flowing over the Internet is composed of visual data, images and videos.
This wealth of data is a precious source of information for a number of critical applications.
In the wake of recent acts of terrorism, for example,
investigators relied heavily on videos posted on social networks by a multitude of individual users.
However, armed with sophisticated video editing tools,
a skilled attacker can prepare and diffuse false videos that may fool even expert analysts, and may cause huge damage.
In this context, like in many others, establishing the integrity of visual assets of interest
becomes of primary importance \cite{Milani2012}.

Recently, there has been intense activity on video forgery detection and localization.
These papers, however, focused only on a few specific attacks,
trying to detect frame erasure/substitution \cite{Stamm2012, Gironi2014},
compression artifacts \cite{Wang2006, Jiang2013, Labartino2013}, or video object copy-move \cite{Bestagini2013, DAmiano2015}.
Therefore, it is important to devise new tools that are able to expose different types of video forgery attacks,
even in the typically adverse conditions encountered on social networks.
In this work we will focus on splicings carried out with chroma-key composition (see Fig.1)
and adopt an anomaly detection strategy.
The forgery is singled out based on the statistical differences, observed in a suitable feature domain, with respect to genuine material.

To implement this idea we resort to deep learning tools, including auto-encoders and recurrent neural networks.
A major qualifying point is the use of unsupervised learning:
network parameters are learnt based on the very same video under analysis.
However, we assume that a small number of frames in the video under test is known to be pristine,
an hypothesis easily satisfied by background-only segments.
The proposed method uses residual-based features, as done in \cite{Verdoliva2014, Cozzolino2015},
which are processed by autoencoders in combination with LSTM recurrent neural networks.
During the training phase,
the autoencoder learns to reproduce faithfully the pristine input by retaining all relevant information in the hidden layer.
Then, in normal operations, it keeps reconstructing the input with high accuracy, as long as it comes from the same source (the pristine video).
In the presence of spliced areas, the reconstruction error increases significantly, triggering detection.
Therefore, the autoencoder behaves as a one-class classifier:
a large reconstruction error between input and output is interpreted as an anomaly, hence a forgery.
The reconstruction error, computed in sliding window modality over the frame,
is converted into a heat map for visual inspection and, after thresholding, in the final detection map.
The use of recurrent networks, then, allows us to gather information from the whole video, not just a single frame, thus exploiting temporal dependencies.
Autoencoders have been recently used also in \cite{Cozzolino2016},
but in the context of an iterative procedure,
in conjunction with a discriminative labeling step, for single-image forgery detection.
It is also worth noting that in the literature autoencoders have been already proposed
for anomaly detection in order to solve a one class classification problem \cite{Hawkins2002},
and have been applied in very different applications \cite{Marchi2015, Hasan2016}.

\begin{figure}[t!]
	\centering\footnotesize
	\setlength{\tabcolsep}{1pt}
	\begin{tabular}{cccc}
		\includegraphics[width=.11\textwidth]{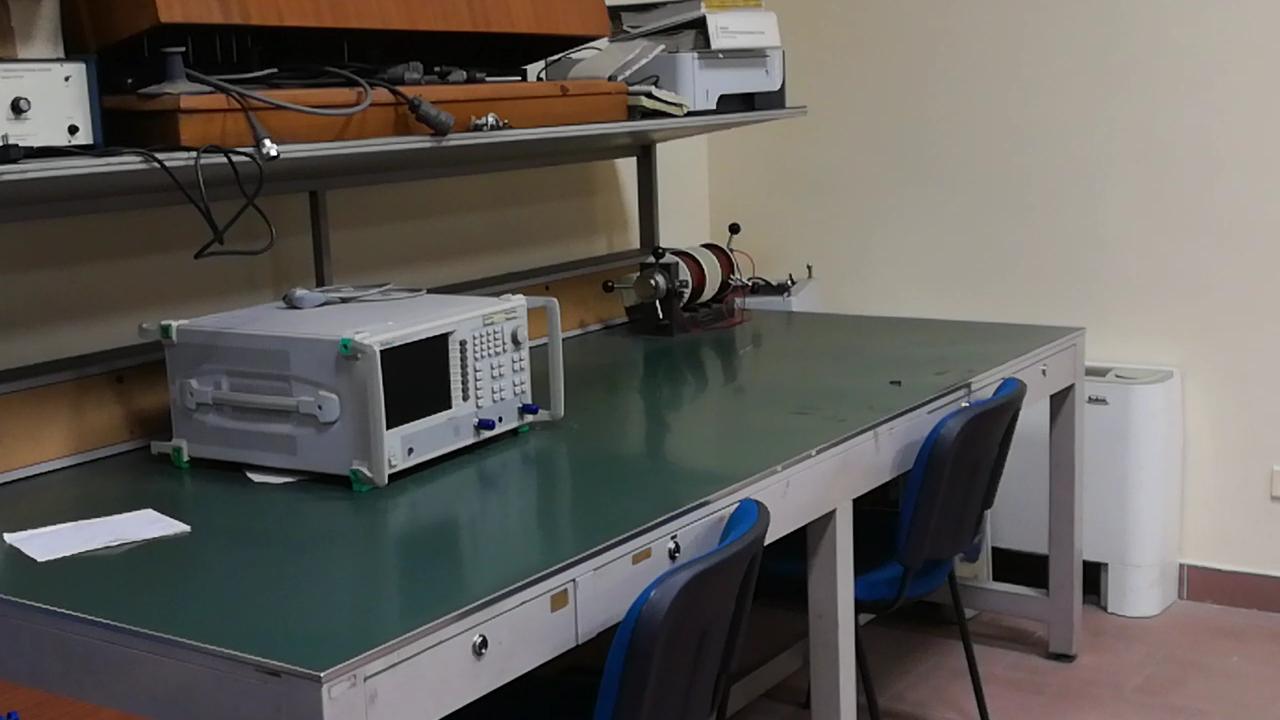} &
		\includegraphics[width=.11\textwidth]{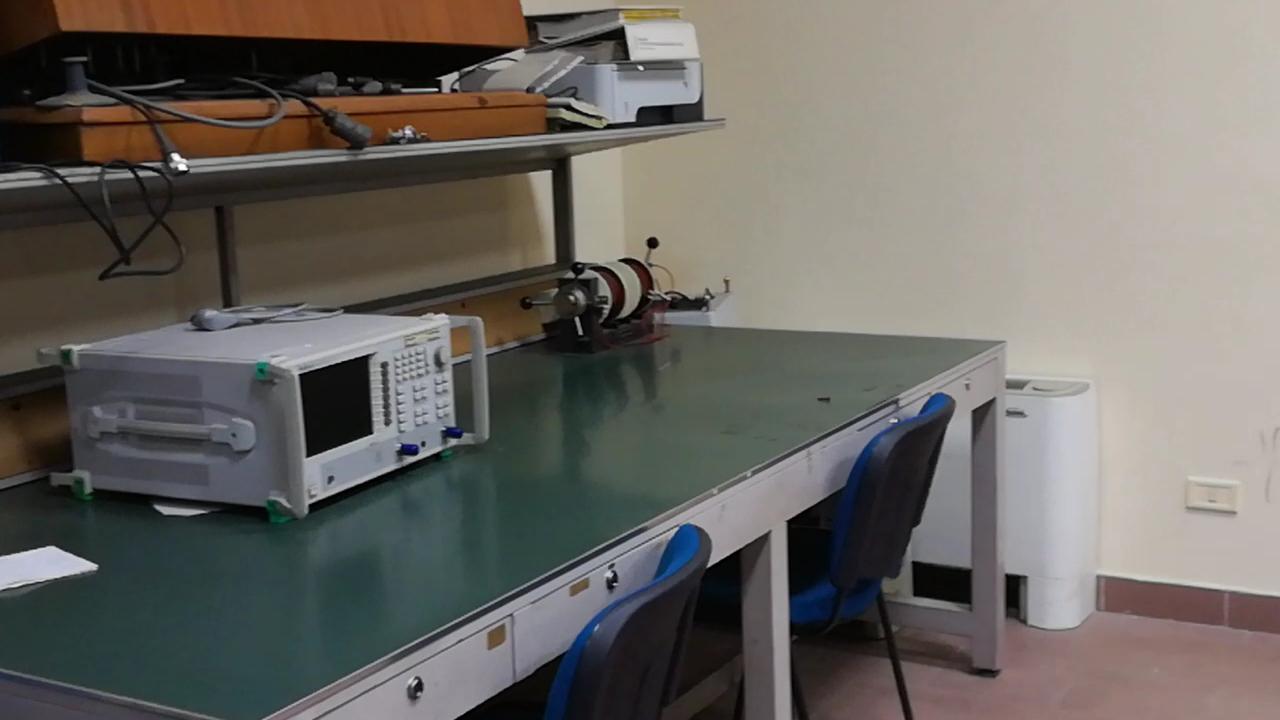} &
		\includegraphics[width=.11\textwidth]{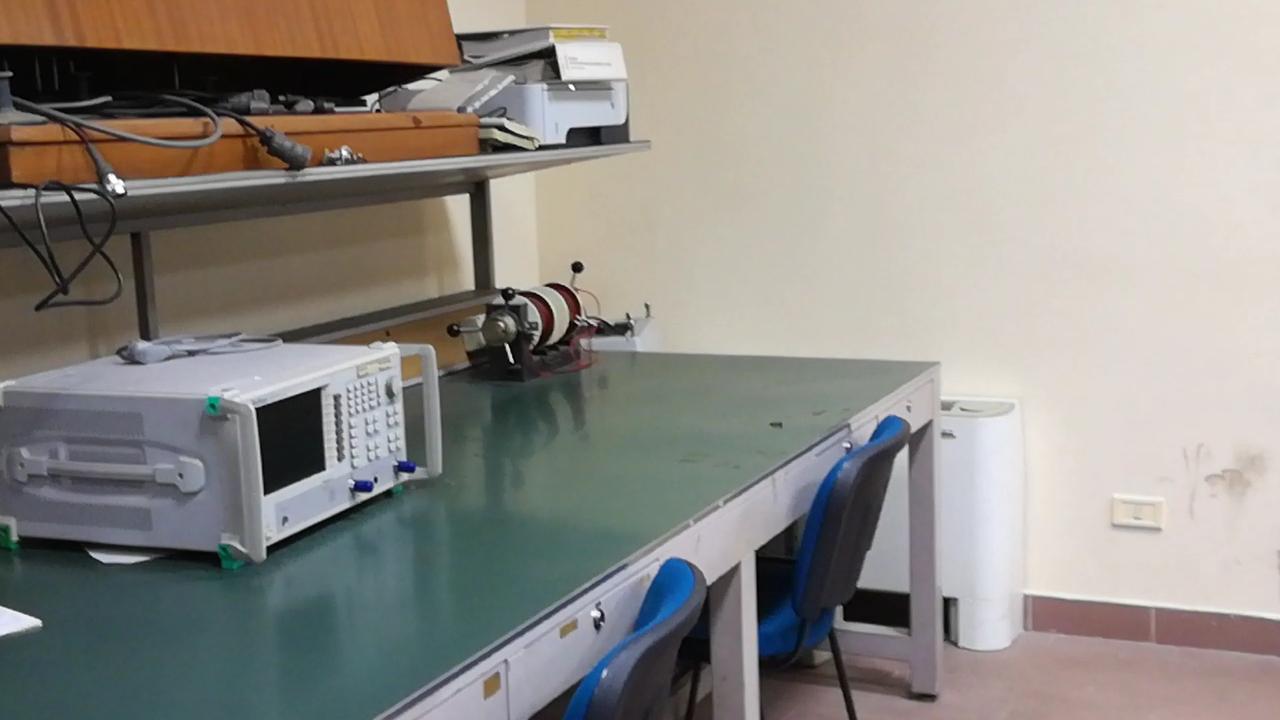} &
		\includegraphics[width=.11\textwidth]{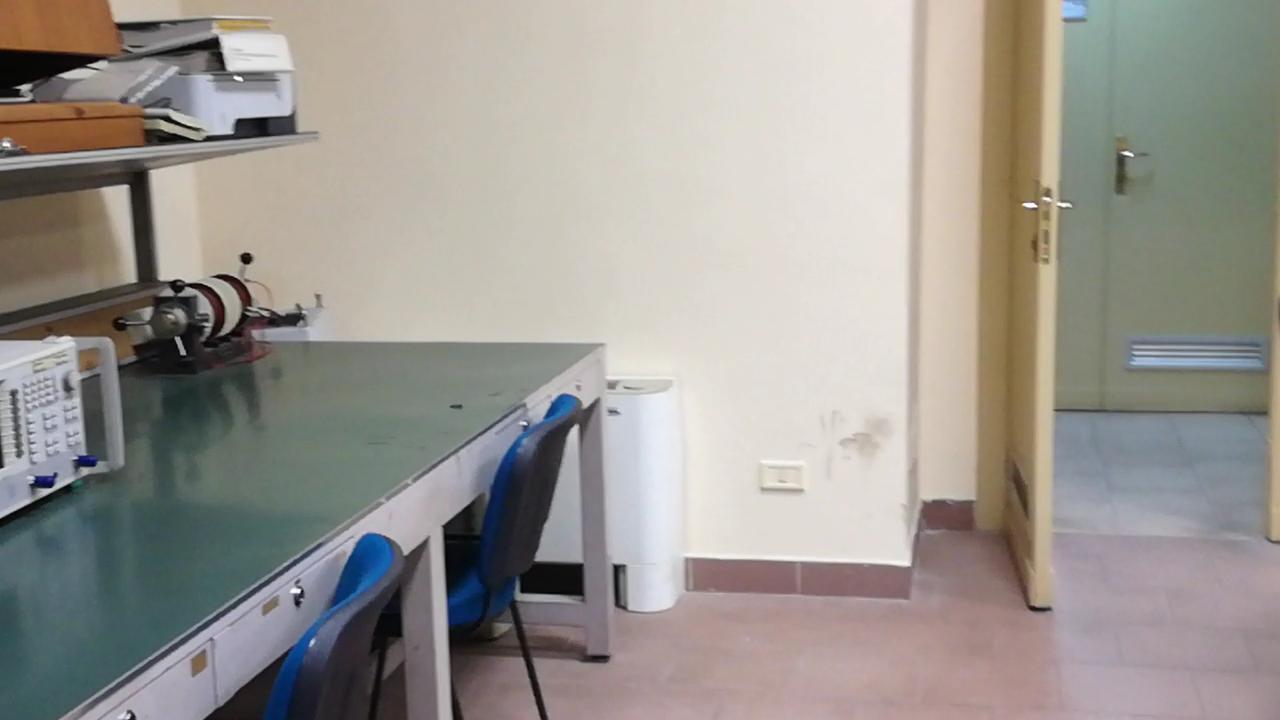} \\

		\includegraphics[width=.11\textwidth]{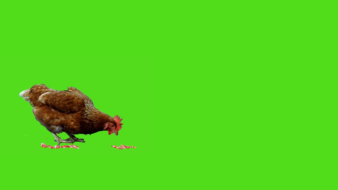} &
		\includegraphics[width=.11\textwidth]{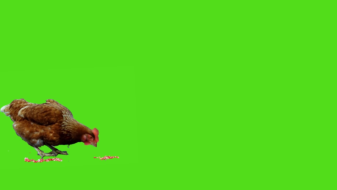} &
		\includegraphics[width=.11\textwidth]{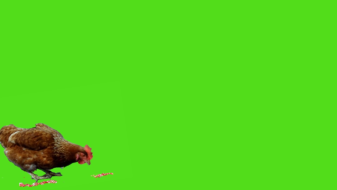} &
		\includegraphics[width=.11\textwidth]{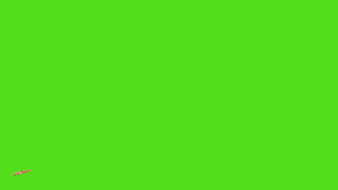} \\

		\includegraphics[width=.11\textwidth]{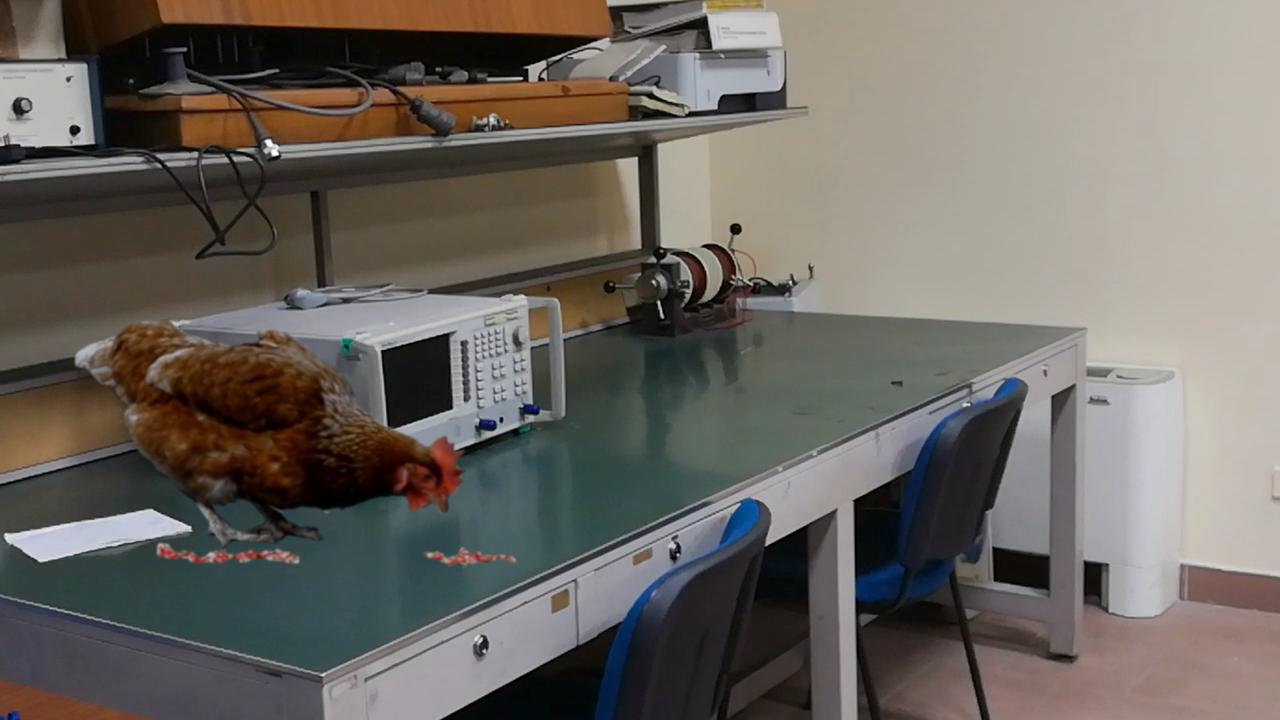} &
		\includegraphics[width=.11\textwidth]{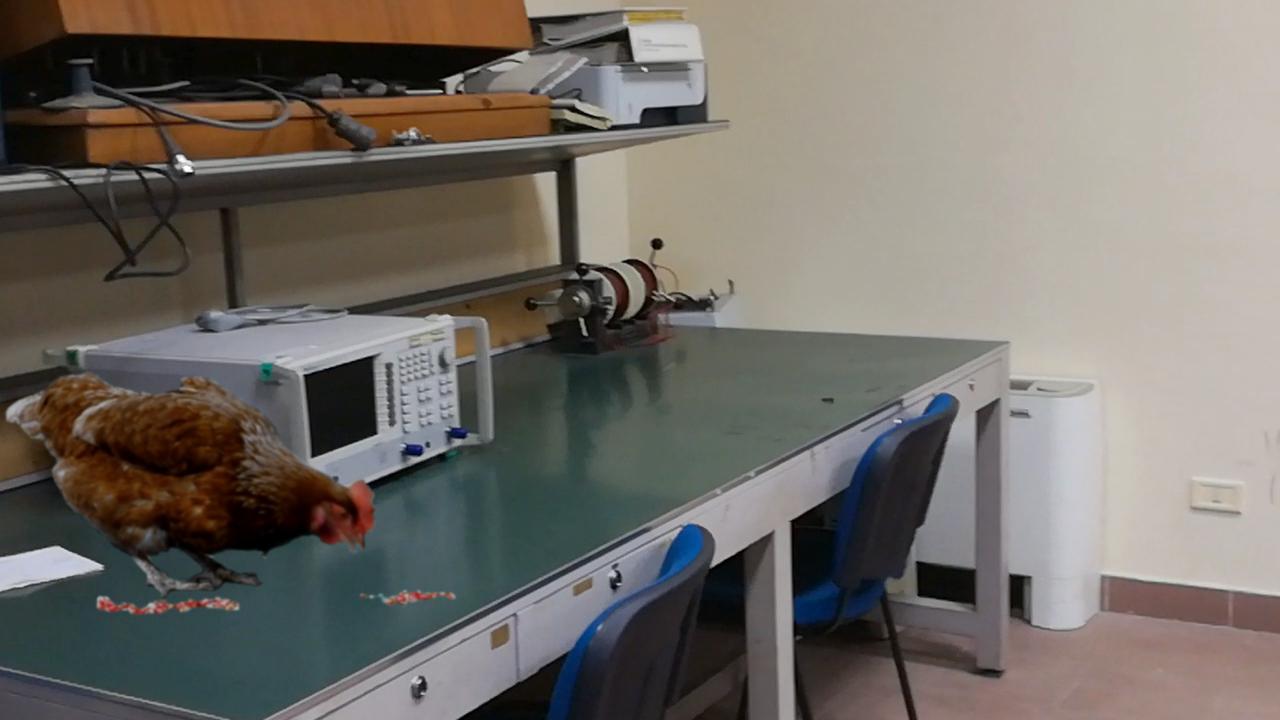} &
		\includegraphics[width=.11\textwidth]{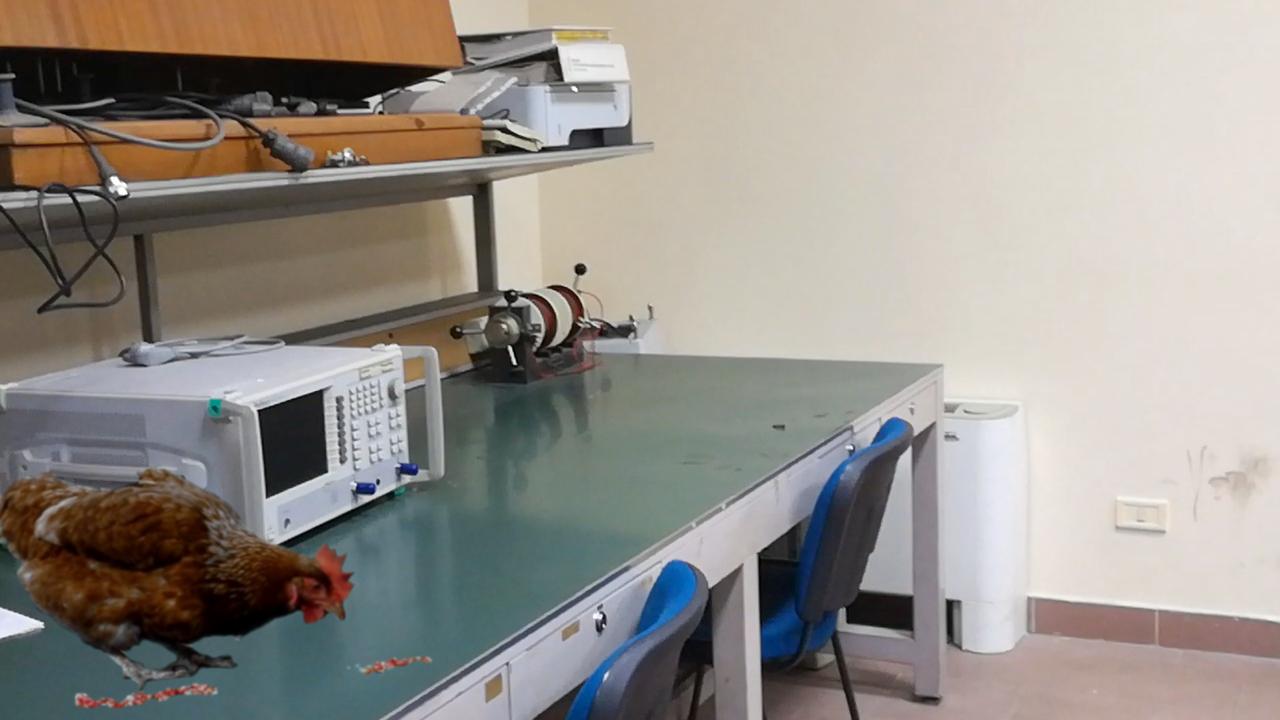} &
		\includegraphics[width=.11\textwidth]{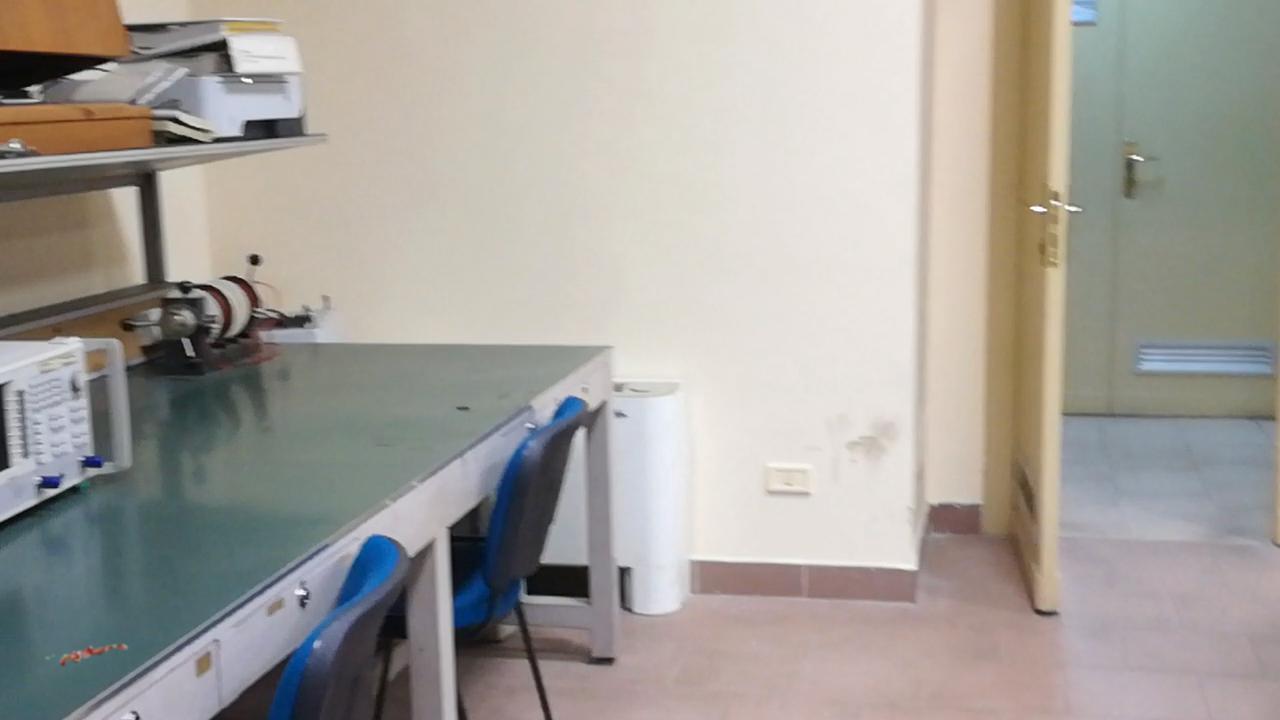} \\
	\end{tabular}
    \vspace{2mm}
	\caption{Example of chroma-key composition on a video sequence. Top: frames from the original video;
		     middle: frames from the aligned green screen video; bottom: forged frames.}
	\label{fig:Example}
\end{figure}

In the rest of the paper we report on related work concerning video forgery detection,
provide some background on the neural networks of interest,
describe the proposed method,
and finally comment the results of some experiments on a suitable dataset of forged videos.

\section{Related Work}
\label{sec:rel-work}

In this section we will briefly review the main research lines
emerging in the recent literature on video forgery detection and localization.
We will focus especially on methods that do not make any specific hypotheses on the type of manipulation.
Then, we will describe techniques that look for splicing performed by croma-key compositing.

A good number of forgery detection methods rely on the camera sensor noise, or Photo Response Non Uniformity (PRNU).
This is a fixed pattern, caused by inhomogeneity in silicon wafers and imperfections in the sensor manufacturing process,
which can be retrieved in all images/videos taken by a given camera.
Uniqueness and stability make of PRNU a sort of camera fingerprint, and a valuable tool for image forensics.
It is a powerful tool for source identification \cite{Lukas2006}, but also for image forgery detection
\cite{Chen2008, Chierchia2014},
since it can be applied irrespective of the specific type of manipulation.
Any tampering of the source data causes also a change in the sensor fingerprint, which raises an alarm on the authenticity of the asset.
Applications to video source identification have been explored originally in \cite{Chen2007},
and subsequently in \cite{Chuang2011}, where a strategy to face the effects of strong compression is proposed.
Instead, its use for forgery detection is first analyzed in \cite{Mondaini2007}.
It is worth noting that PRNU-based forgery detection is much more challenging than source identification,
since the detector must operate on small local patches to locate the forgery.
In this case, the very low intensity of the PRNU pattern becomes a serious limit.
Moreover, the performance is very sensitive to the scene content, limiting application in dark or textured areas \cite{Chen2008}.
A final drawback, is the need of a large training set to estimate the PRNU pattern.

\begin{figure}[t!]
	\centerline{
		\includegraphics[scale=0.9]{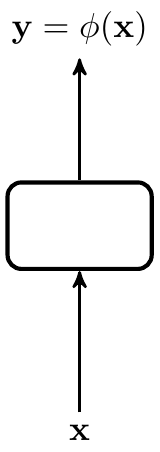} \hspace{6mm}
		\includegraphics[scale=0.9]{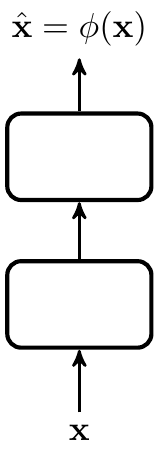}
		\includegraphics[scale=0.9]{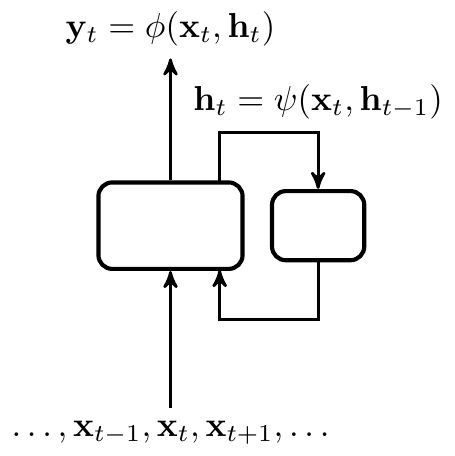}}
	\vspace{3mm}
	\caption{High-level differences among feedforward networks (left), autoencoders (center) and recurrent networks (right).}
	\label{fig:ANN_models}
\end{figure}

Other methods rely on the analysis of noise residuals
obtained by removing the semantic content of the video through high-pass filtering \cite{Pandey2016}.
In fact, noise residuals bear traces of peculiar in-camera processes which can be used to detect statistical anomalies.
Unlike PRNU-based methods, these do not require training data, and work directly on the residuals frames.
In \cite{Hsu2008} a Gaussian mixture model is used to analyze the temporal correlation on residuals.
However, the method is very sensitive to quantization noise and cannot handle videos with moving background.
This last restriction applies also to the method proposed in \cite{Kobayashi2010},
where the inconsistencies of the photon shot noise characteristics are used for forgery detection.
Chen et al. \cite{Chen2016} focus instead on motion residuals, looking for irregularities.
However, the proposed method can only identify the forged video segments,
without localizing the specific forged area.

We conclude this short survey of methods focusing on the detection of splicing carried out through chroma-key compositing.
A few papers in the literature deal with this problem.
They all try to detect artifacts generated by the foreground/background composition, relying on their statistical differences.
Su et al. \cite{Su2011} look for different correlation patterns, in the green component, around the edges of the spliced area.
In \cite{Xu2012}, instead, inconsistencies of statistical features of quantized DCT coefficients are sought, while in \cite{Bajiwa2016} blurring artifacts are considered.

\section{Background}
\label{sec:background}

In this section, we provide some basic information about the artificial neural network models adopted in this work, that is, autoencoders, and recurrent neural networks.
Fig.\ref{fig:ANN_models} provides a very high-level summary of the major differences among these models.
A generic feedforward neural net (on the left) can compute any function $\y=\phi(\x)$ of the input vector.
The input vector has a fixed length, corresponding to the size of the input layer, while no constraint is imposed on the output.
Autoencoders (center) are feeforward neural networks in which the output is constrained to have the same size as the input.
Indeed, they are usually trained to provide as output the best possible approximation of the input, that is $\y=\wx=\phi(\x)$.
Recurrent neural networks (right) have been introduced to deal with input data coming in sequential form,
such as the words of a sentence or the frames of a video.
In such cases, it is often desired that the output at a given time instant, $\y_t$,
depends not just on the input at the same instant, $\x_t$, but also on past inputs, $\x_0, \ldots, \x_{t-1}$.
This can be obtained efficiently by introducing a state variable, ${\bf h}_t$, which keeps memory of past inputs by a suitable updating rule,
obtaining the model depicted in Fig.\ref{fig:ANN_models}, with $\y_t=\phi(\x_t,{\bf h}_t)$, and ${\bf h}_t=\psi(\x_t,{\bf h}_{t-1})$.
In the following, we describe in some more detail these networks.

\begin{figure}[t!]
	\centerline{\includegraphics[scale=0.37]{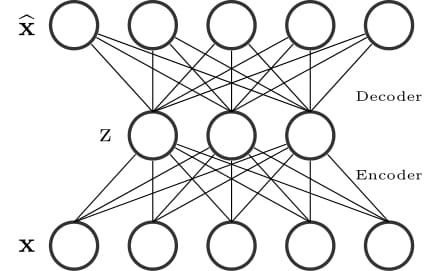}}
	\vspace{9mm}
	\caption{An autoencoder with a single hidden layer.}
	\label{fig:autoencoder}
\end{figure}

\subsection{Autoencoders}

An autoencoder is able to learn, by proper training, a representation (coding)
of the input data in oder to guarantee some desired properties \cite{Bengio2013}.
Its simplest form is shown in Fig.3: a feedforward non-recurrent net
with the input layer connected to the output layer through a single hidden layer.
Typically, the input and output layers have the same size, and the output is desired to approximate the input.
In this case, one can evaluate the reconstruction error between input and output by means of a specific loss function.
Specifically,
the encoder maps the input vector, $\x \in \RR^K$, to its hidden (or latent) representation, $\z \in \RR^H$, as
\begin{equation}
\z  = \phi_1 \left( \bf{A}_1 \x + \bf{b}_1 \right)
\end{equation}
where $\phi_1(\cdot)$ is the activation function, $\bf{A}_1$ is a $K\times H$ weight matrix and $\bf{b}_1$ is a bias vector.
The decoder, instead, maps the hidden representation to the output reconstruction as
\begin{equation}
\wx = \phi_2 \left( \bf{A}_2 \z + \bf{b}_2 \right)
\end{equation}
where $\phi_2, \bf{A}_2$ and $\bf{b}_2$ are the activation function , the weight matrix and the bias vector, respectively.
All these parameters $\theta = \{{\bf{A_1, b_1, A_2, b_2}}\}$ are learned
by minimizing the average reconstruction error between input and output over a suitable training set.
Usually, the average Euclidean distance is used as loss function:
\begin{equation}
L(\theta) =  ||\x - \wx; \theta||^2
\end{equation}
and the optimization is carried out through stochastic gradient descent.

Most of the times, autoencoders have a bottleneck structure, $H<K$, aimed at extracting a compact representation of the input.
In this way, the network is forced to represent the input with a smaller number of variables (in the hidden layer)
while preserving the information content as much as possible.
If linear activation functions are used, the autoencoder approximates a principal component analysis (PCA),
providing a low-dimensional linear representation of data.
More often, nonlinear activation functions are used, such as rectified linear unit (ReLU), hyperbolic tangent, or a sigmoid function.
In this case the autoencoder goes beyond PCA, capturing multi-modal aspects of the input distribution \cite{Japkowicz2000}.

\subsection{Recurrent Neural Networks}
\label{sec:rnn}
In feedforward neural networks, information flows only from input units to output units.
Every time the network is fed with a new input, all previous ones are forgotten.
However, when dealing with sequential data, such as audio signals, sentences, videos, and so on,
it is very likely that previous inputs carry valuable information on the current task,
and hence the network should be able to \emph{remember} as many inputs as possible, and take them into account.

To address this problem, one can resort to recurrent neural networks (RNN)
which receive a sequence of values in input, $\x_{0}, \x_{1}, \ldots, \x_{t}, \ldots$,
and produce a corresponding sequence of output values, $\y_{0}, \y_{1}, \ldots, \y_{t}, \ldots$
In principle, each output value $\y_t$ depends on all past inputs.
In practice, the network maintains a hidden state ${\bf h}_t$, which represents the \emph{memory} of the network.
State and output are updated with each new input
\begin{eqnarray}
\mathbf{h}_t & = & \sigma (\mathbf{Wx}_{t}+\mathbf{Uh}_{t-1}) \\
\mathbf{y}_t & = & \sigma (\mathbf{Vh}_{t})
\label{eq:rnn-output}
\end{eqnarray}
where $\sigma$ indicated a sigmoid nonlinearity, and all network parameters ${\bf W,U,V}$ must be learned by suitable training.

For training, feedforward networks rely on the well-known backpropagation algorithm,
together with an optimization method such as the gradient descent.
The same approach can be applied to RNNs, obtaining a training algorithm known as backpropagation through time (BPTT), where the output error backpropagates across all timesteps.
Unfortunately, the vanishing-gradient effect, which already affects feedforward networks, becomes a major problem with RNNs.
The error, propagating through a large number of timesteps, tends to vanish, preventing the network from learning long-term dependencies.

A number of solutions to this problem have been proposed, with mixed results
\cite{Pascanu2013}.
Recently, the long short-term memory (LSTM) model has emerged as the most promising architecture for effective training.
As the name suggests, a LSTM network is able to learn both short-term and long-term dependencies, thanks to the gate-based mechanism shown in Fig.4.
There are three gates, the forget, input, and output gates, which control, respectively,
the fraction of the current state that should be discarded, the fraction of the current input that should be used to update the state, and the part of the newly computed state to use for updating the output.
While the network becomes more complex,
it also gains the ability to select which parts of the current and past information to use for computing the output and updating the state.

\begin{figure}[!t]
	\centerline{\includegraphics[width=6.5cm]{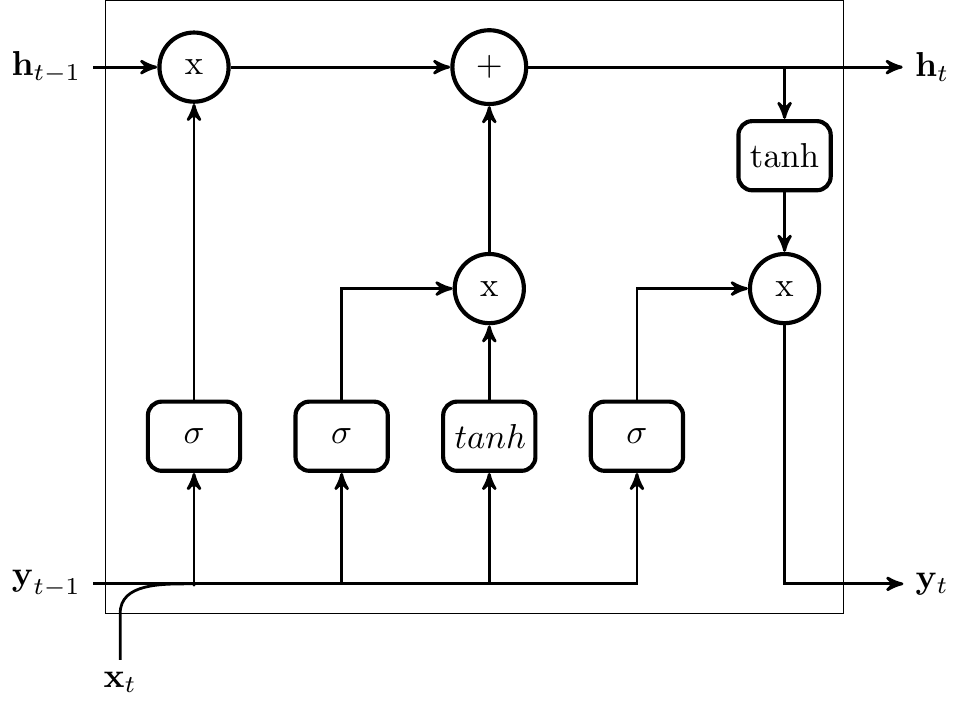}}
    \vspace{2mm}
	\caption{LSTM cell.}
	\label{fig:lstm-gates}
\end{figure}

\section{Proposed method}

The method proposed here takes inspiration from our
previous work on the detection and localization of splicing in still images
\cite{Cozzolino2015, Cozzolino2016}.
We provide, first, a brief overview of the method, and then describe in more detail its components.

Input frames are analyzed in sliding window modality, using image patches of 128$\times$128 pixels taken with stride 8.
From each patch, we extract a handcrafted feature,
designed to capture the subtle statistical differences of spliced material with respect to the original video.
Each feature is analyzed by an anomaly detector, which outputs an anomaly score.
All such scores are then projected back on the image domain and aggregated to produce a global ``heat'' map
which is used as the basis to detect the spliced material, if present, and its position.
Note that we work on single-frame 2d patches.
However, temporal dependencies are taken into account through the use of recurrent neural networks.

In the following subsections we describe the features and the anomaly detector, the latter based on a recurrent autoencoder.

\subsection{Feature extraction}

The literature on image forgery detection makes very clear the importance
of using features extracted from image residuals \cite{Pandey2016, Cozzolino2014a, Li2016},
that is, after removing the semantic image content, which does not help discovering statistical anomalies.
In particular, high-order statistics seem necessary to achieve a good discrimination \cite{Verdoliva2014, Cozzolino2015},
which can be considered through co-occurrence counts.
Here, we use features inspired by work in steganalysis \cite{Pevny2010, Fridrich2012}.
However, unlike in steganalysis,
where a rich set of different high-pass linear and non-linear residuals seems to be necessary to achieve a good performance,
we consider a single high-pass third-order derivative filter to extract meaningful features.
This choice proved effective in previous work \cite{Cozzolino2014a},
as also confirmed by recent experiments carried out in \cite{Li2016} on a variety of different manipulations.

We therefore follow a three-step model comprising
\begin{enumerate}
	\item   computation of residuals through high-pass filtering;
	\item   quantization of the residuals;
	\item   computation of a histogram of co-occurrences.
\end{enumerate}
The final histogram is the feature vector associated with the whole patch, to be used for anomaly detection.
The residual is obtained as $r_{ij} = f_{i,j-1} -3\,f_{i,j} +3\,f_{i,j+1} -\,f_{i,j+2}$,
where $f$ and $r$ are origin and residual frames, and $i,j$ indicate spatial coordinates.
However, we are interested in the {\em joint} distribution of residuals,
which provides information on higher-order phenomena and involves a larger neighborhood.
To obtain a manageable histogram of co-occurrences, residuals are therefore quantized to a few bins and truncated,
$\hr_{i,j} = {\rm trunc}_T({\rm round}(r_{i,j}/q))$,
with $q$ the quantization step and ${\rm trunc}_T$ the truncation operator at level $T$.
In particular, we use a uniform five-level quantizer, and compute co-occurrences on four pixels in a row, that is
\begin{align*}
&  C(k_0,k_1,k_2,k_3) = \\
&  \sum_{i,j} I(\hr_{i,j}=k_0,\hr_{i+1,j}=k_1,\hr_{i+2,j}=k_2,\hr_{i+3,j}=k_3)
\end{align*}
where $I(E)$ is the event indicator.
The homologous column-wise co-occurrences are pooled with the above, based on symmetry considerations, obtaining eventually a 625-bin histogram, which is reduced to 338 by further symmetry arguments.
Finally, the histogram is converted to vector form and normalized to zero mean and unit norm to obtain the final feature vector ${\bf x}$.

\begin{figure}[t!]
	\centerline{\includegraphics[width=8cm]{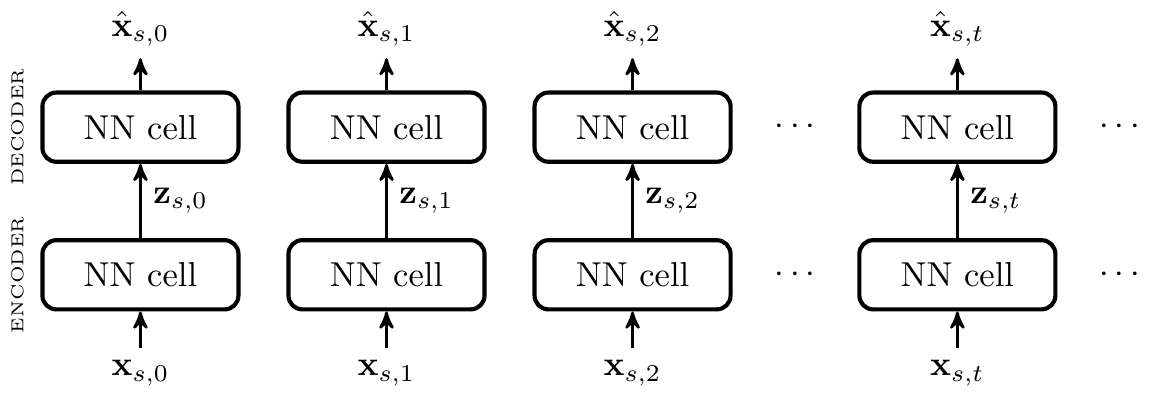}}
	\vspace{2mm}
	\caption{An array of identical feedforward autoencoders for a sequence of input patches.}
	\label{fig:feedforward_AE}
\end{figure}

\subsection{Anomaly detection}

To perform anomaly detection, we exploit the properties of autoencoders.
The parameters of the autoencoder are learned on a dataset of pristine feature vectors, extracted from a short sequence of splicing-free frames.
During the test phase,
whenever a pristine feature vector is presented in input, the network succeeds in reproducing it with a small error.
On the contrary, in the presence of spliced material, the feature vector does not fit the intrinsic model stored in the network parameter,
and is reproduced with a large error.
By measuring the reconstruction error (3) between input and output one obtains therefore a reliable anomaly score.

With reference to the $t-th$ frame of the video,
the same procedure is applied to all feature vectors, $\x_{s,t}$, with $s$ indicating the spatial location of the corresponding patch.
To build the output heat map, all original patches are replaced by the corresponding anomaly scores.

\begin{figure}[t!]
	\centerline{\includegraphics[width=8cm]{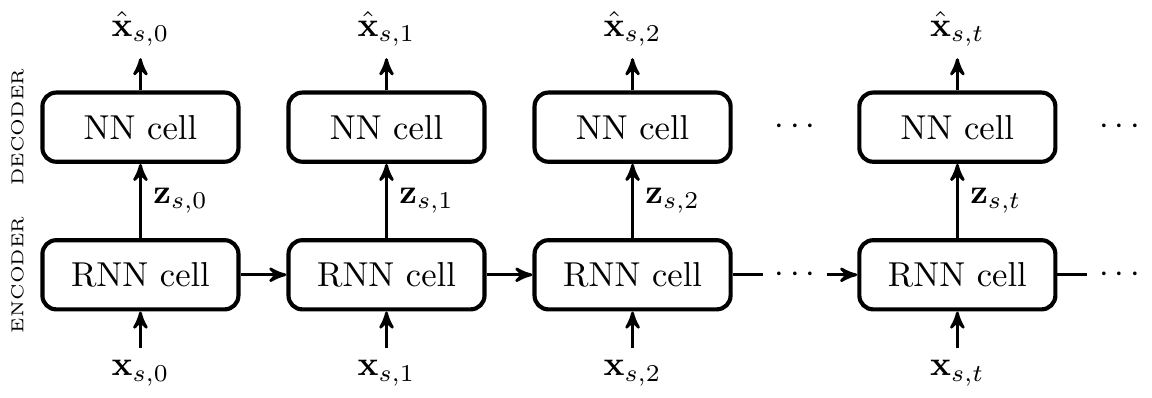}}
    \vspace{2mm}
	\caption{An array of temporally dependent autoencoders for a sequence of input patches.}
	\label{fig:recurrent_AE}
\end{figure}

Under a different perspective,
if we consider all patches associated with a specific spatial location, $s$, they are processed by a virtual array of identical autoencoders,
as shown in Fig.\ref{fig:feedforward_AE}.
However, it can be expected that past input patches carry valuable information on the current one,
which should be taken into account to improve performance.
Therefore, we replace our feedforward autoencoder with a recurrent one, giving rise, after temporal unrolling,
to the structure depicted in Fig.\ref{fig:recurrent_AE}.
The current feature vector is encoded by taking into account also past patches, with a relatively long memory implemented by the LSTM model.
As in the previous structure,
all parameters of this recurrent autoencoder are learnt in advance on a short sequence of pristine frames.

\section{Experimental results}

This section is devoted to the experimental assessment of performance.
We will first present the video dataset,
then analyze the performance of the proposed method,
and finally carry out a comparison with some state-of-the-art reference algorithms.

To assess the performance of the proposed method we created a new specific dataset, available online at www.grip.unina.it, which is described below.
This is primarily due to the scarcity of datasets for video forensics,
but also to the need to perform experiments in perfectly known and controlled conditions,
including knowledge of the origin of videos and of the types of attack.
Our dataset comprises 10 short videos with splicings created with Adobe After Effects CC$^{\mbox{\scriptsize{\textregistered}}}$
using croma-key compositing by means of green screen acquisitions.
The background videos have been captured by the authors using nine different smartphones,
while the green screen acquisitions (including humans, animals, and objects)
have been downloaded from YouTube under the Creative Commons license,
except for one downloaded from: http://www.hollywoodcamerawork.com/green-screen-plates.html.
In order to study also the realistic case of videos downloaded from a social network, typically with some quality impairments,
the dataset includes, together with the original videos (H.264) and the forged videos
(uncompressed AVI),
also their versions uploaded and downloaded from YouTube, at the maximum quality.
Tab.1 reports some synthetic information on the videos, and in particular the total number of frames and the number of forged frames.
All videos were cropped at the same size of 720$\times$1280 pixels.
Fig.7 and Fig.8 show individual frames extracted from each original and, respectively, forged, videos.
It is worth underlining that all videos are characterized by significant motion of either the background or spliced object.

\begin{figure*}[t]
	\centering
	\setlength{\tabcolsep}{1pt}
	\begin{tabular}{ccccc}
		\includegraphics[width=.194\textwidth]{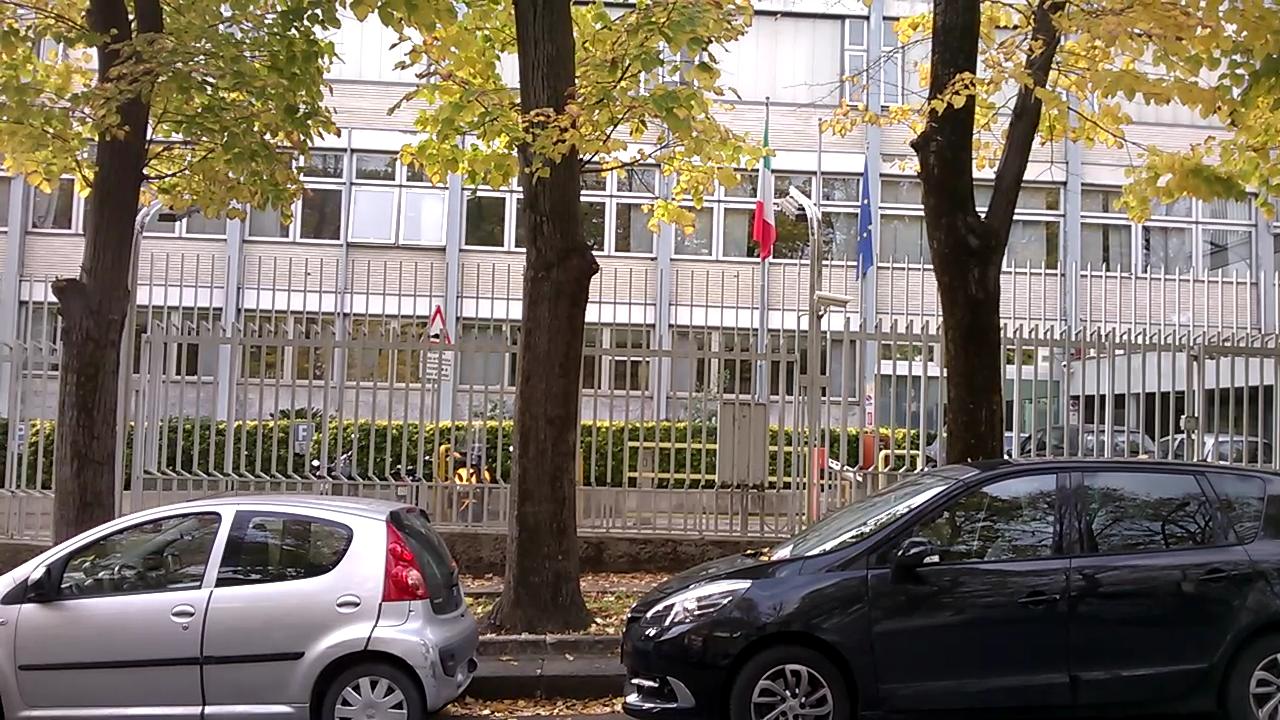} &
		\includegraphics[width=.194\textwidth]{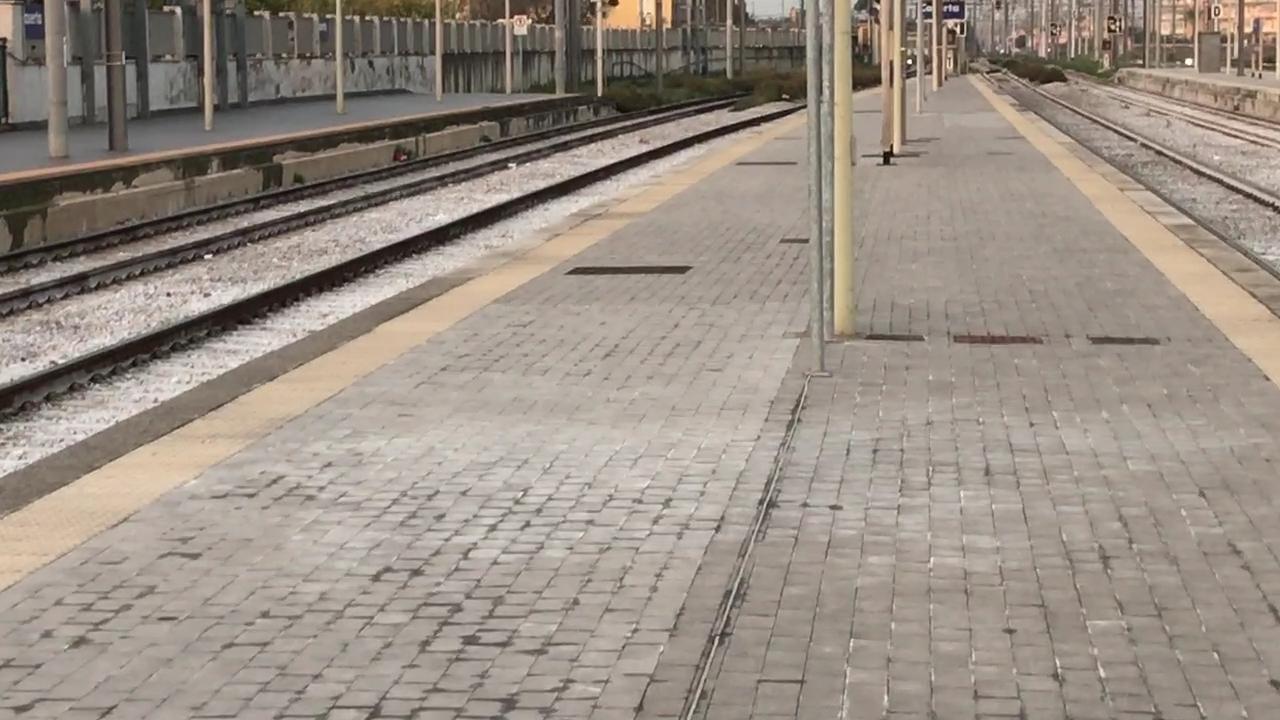}  &
		\includegraphics[width=.194\textwidth]{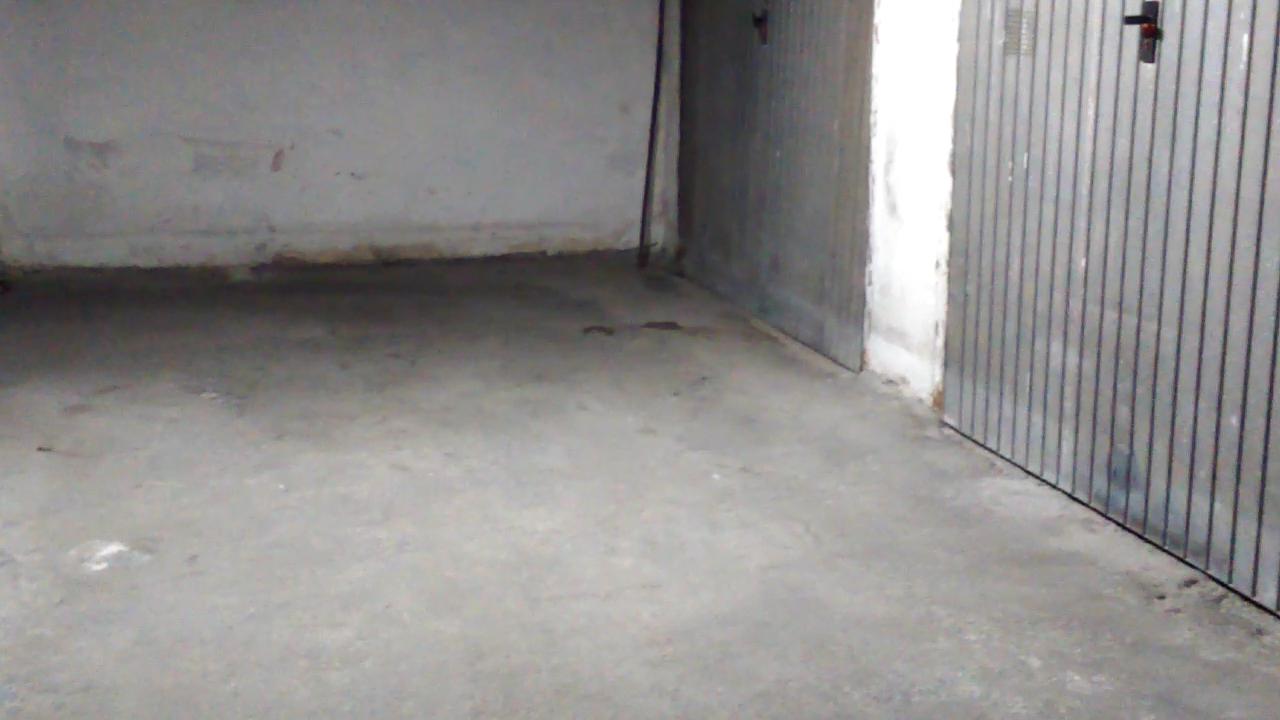} &
		\includegraphics[width=.194\textwidth]{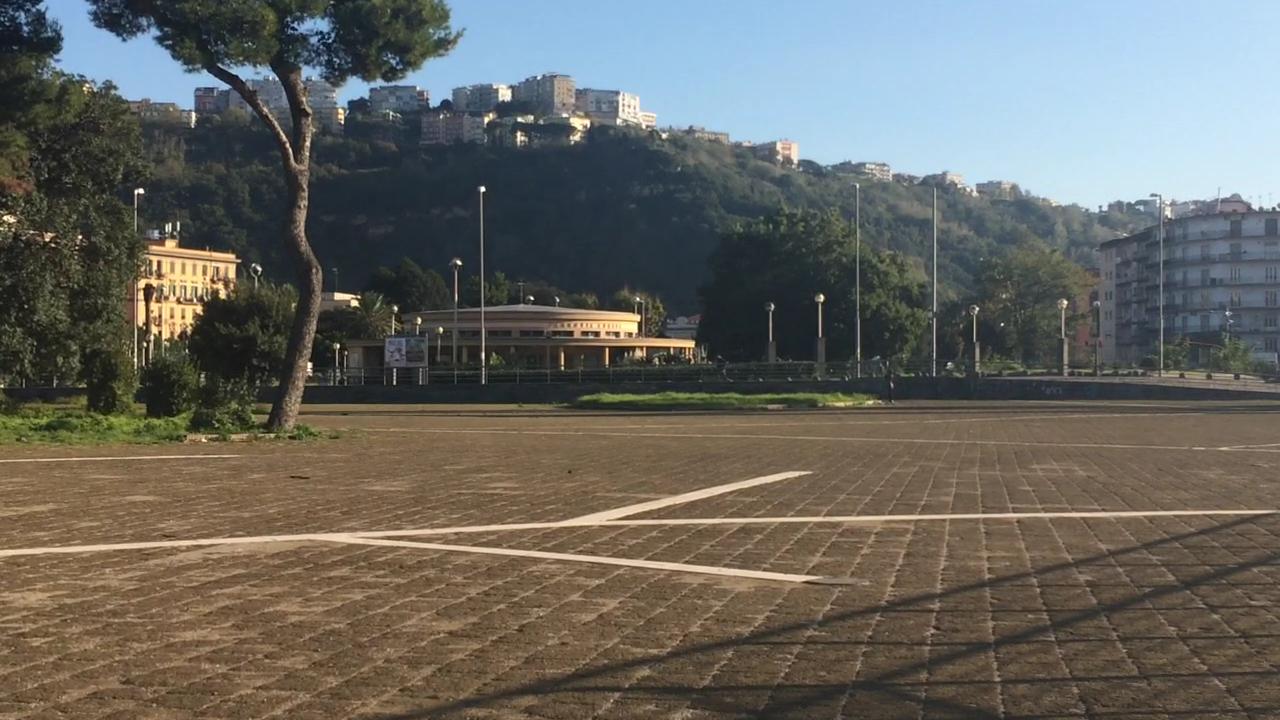} &
		\includegraphics[width=.194\textwidth]{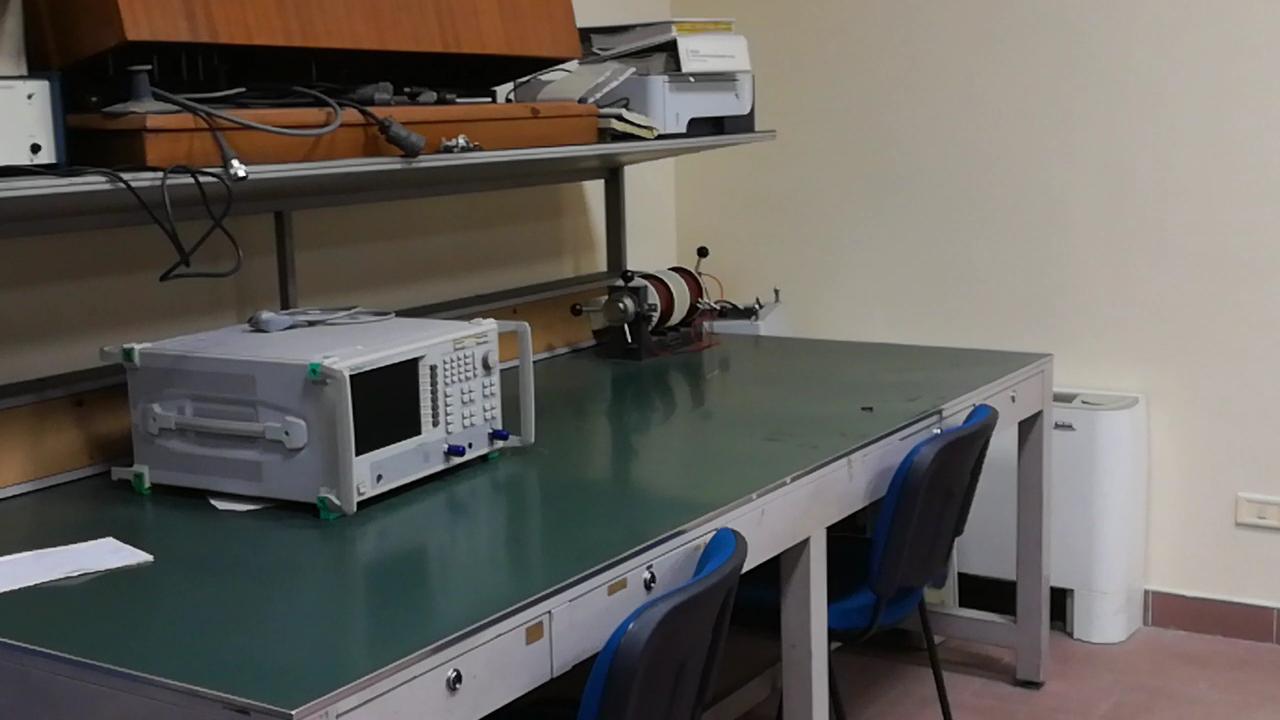}  \\
		\includegraphics[width=.194\textwidth]{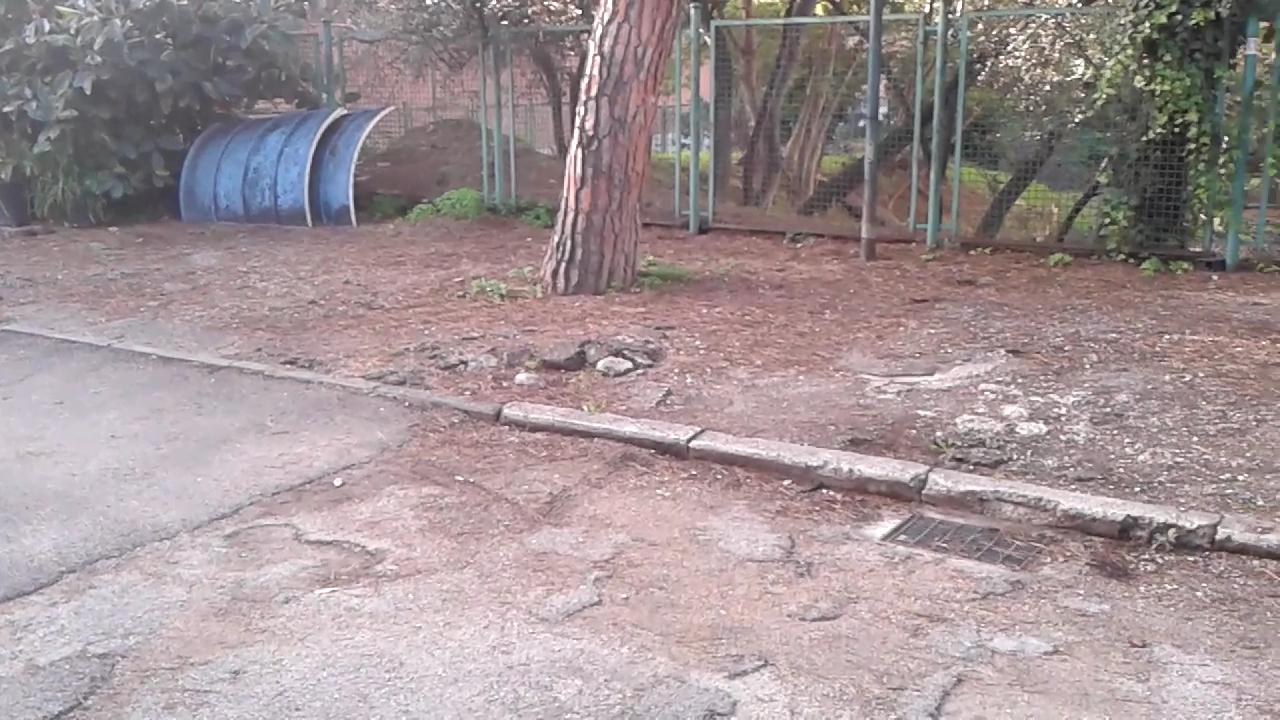} &
		\includegraphics[width=.194\textwidth]{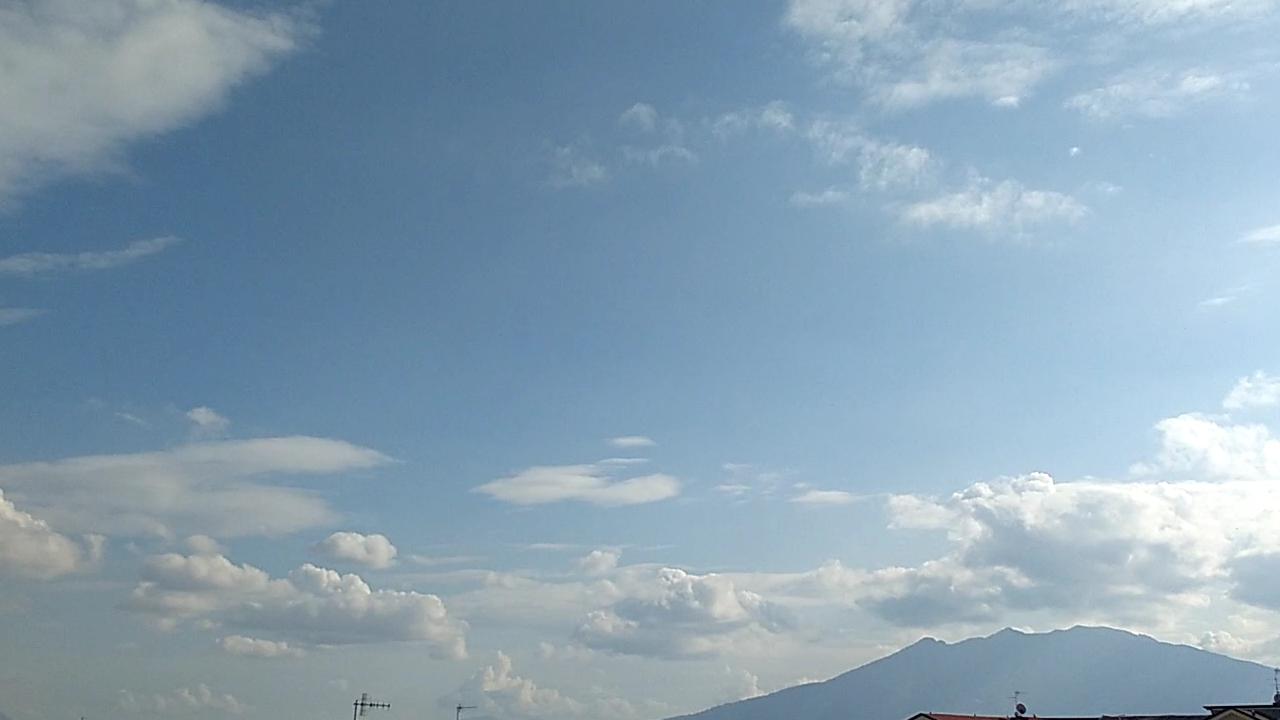} &
		\includegraphics[width=.194\textwidth]{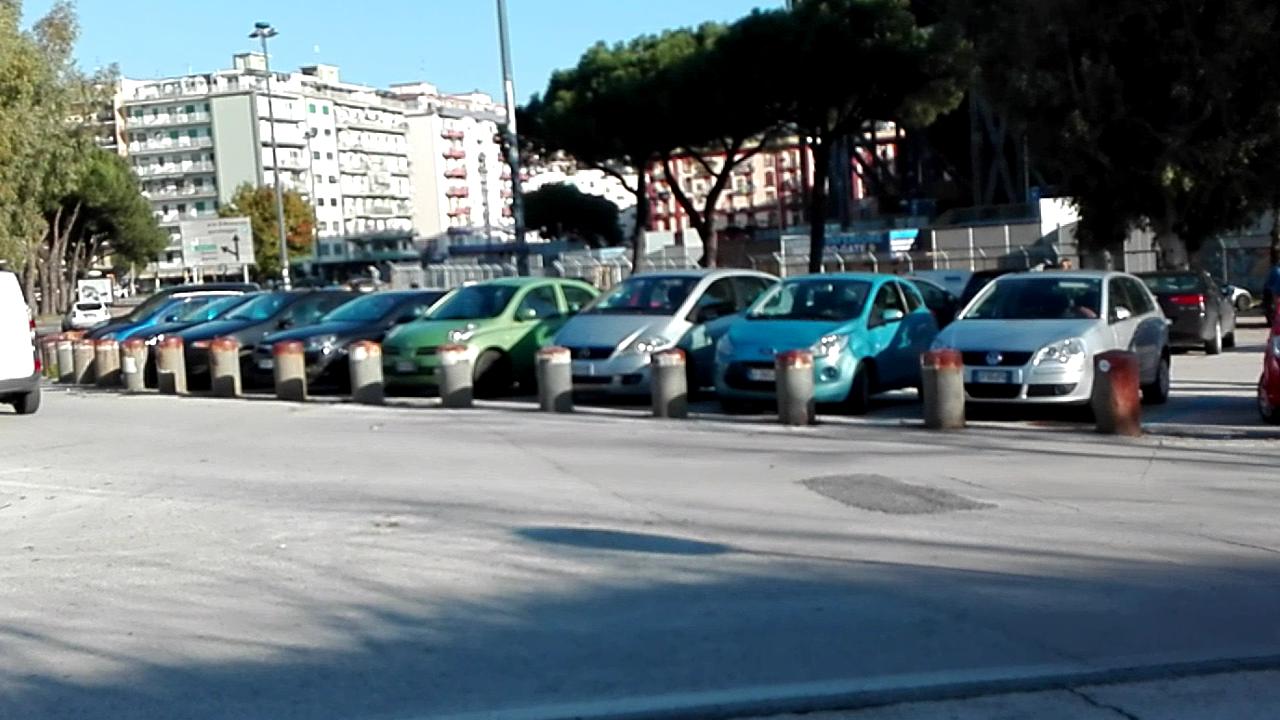}  &
		\includegraphics[width=.194\textwidth]{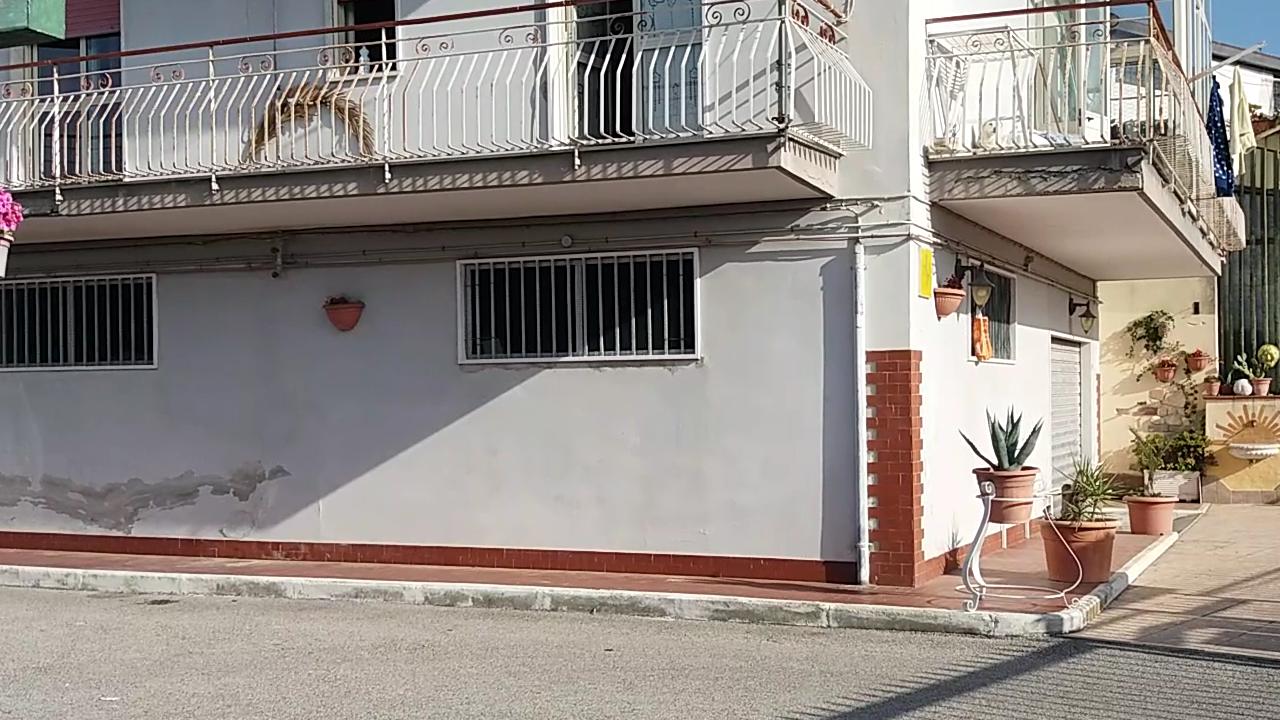} &
		\includegraphics[width=.194\textwidth]{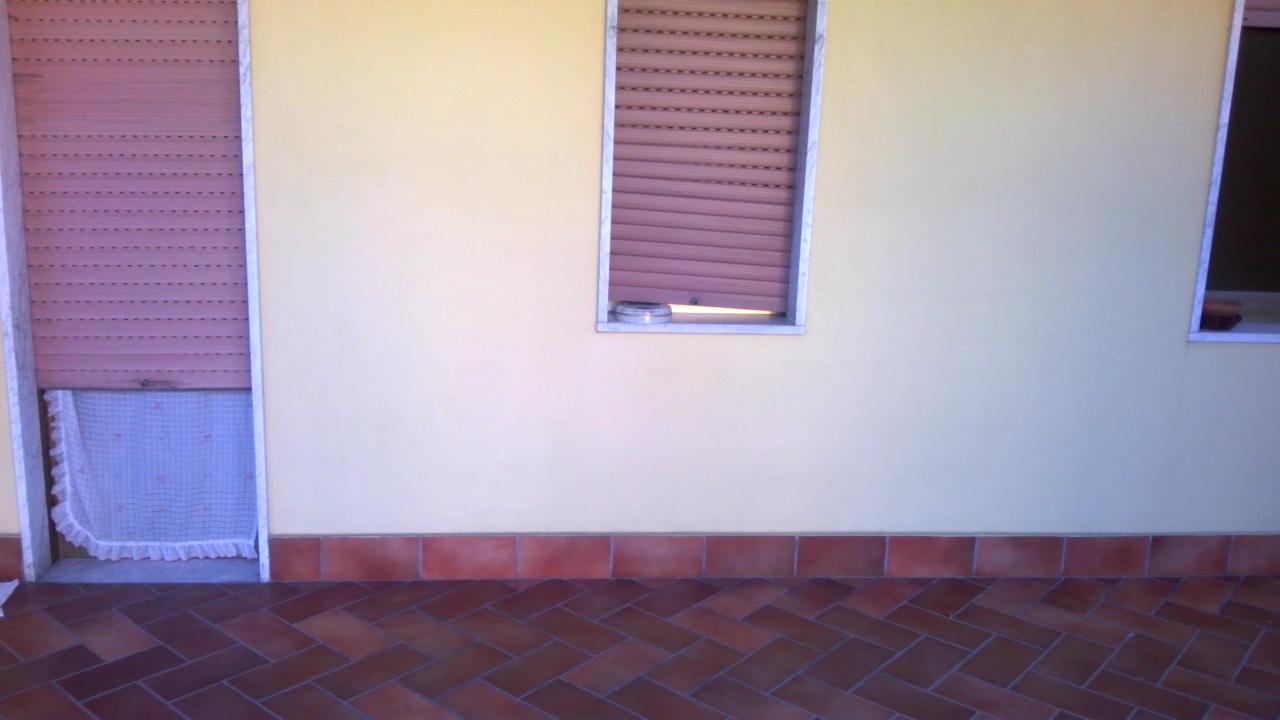} \\
	\end{tabular}
    \vspace{2mm}
    \caption{Individual frames taken from the original videos.}
	\label{fig:dataset}
\end{figure*}

\begin{figure*}[t]
	\centering
	\setlength{\tabcolsep}{1pt}
	\begin{tabular}{ccccc}
		\includegraphics[width=.194\textwidth]{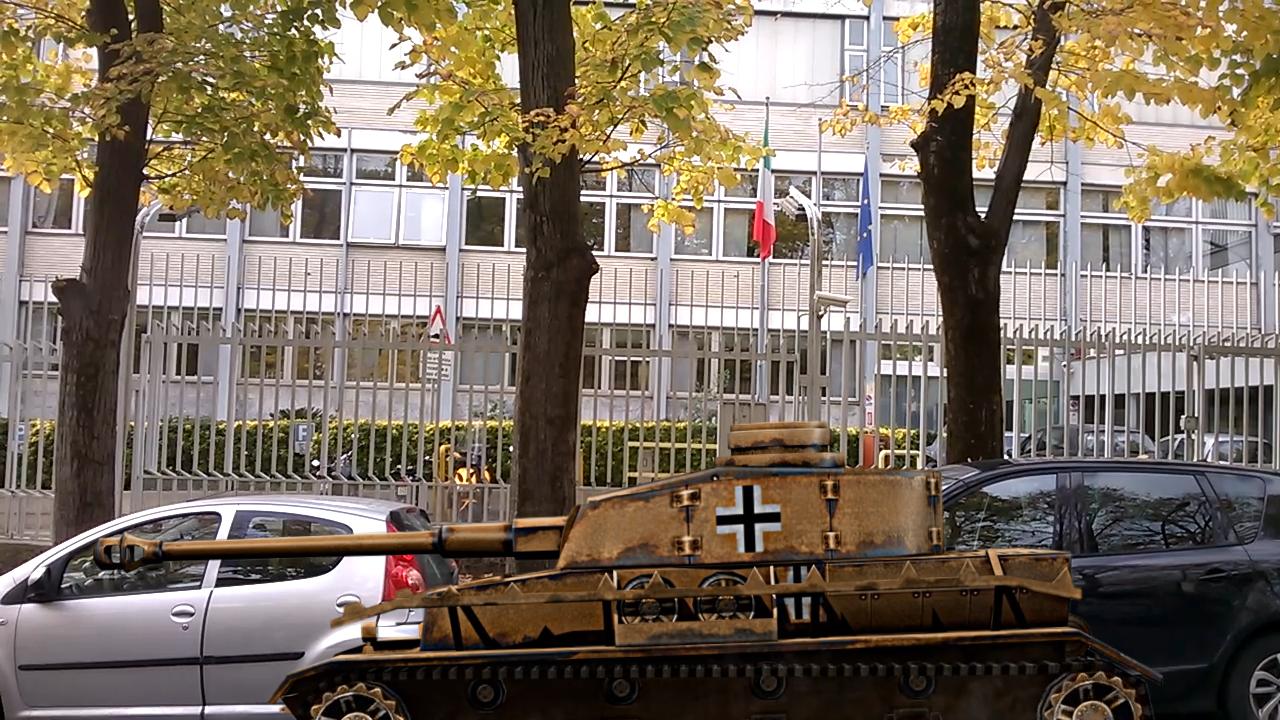} &
		\includegraphics[width=.194\textwidth]{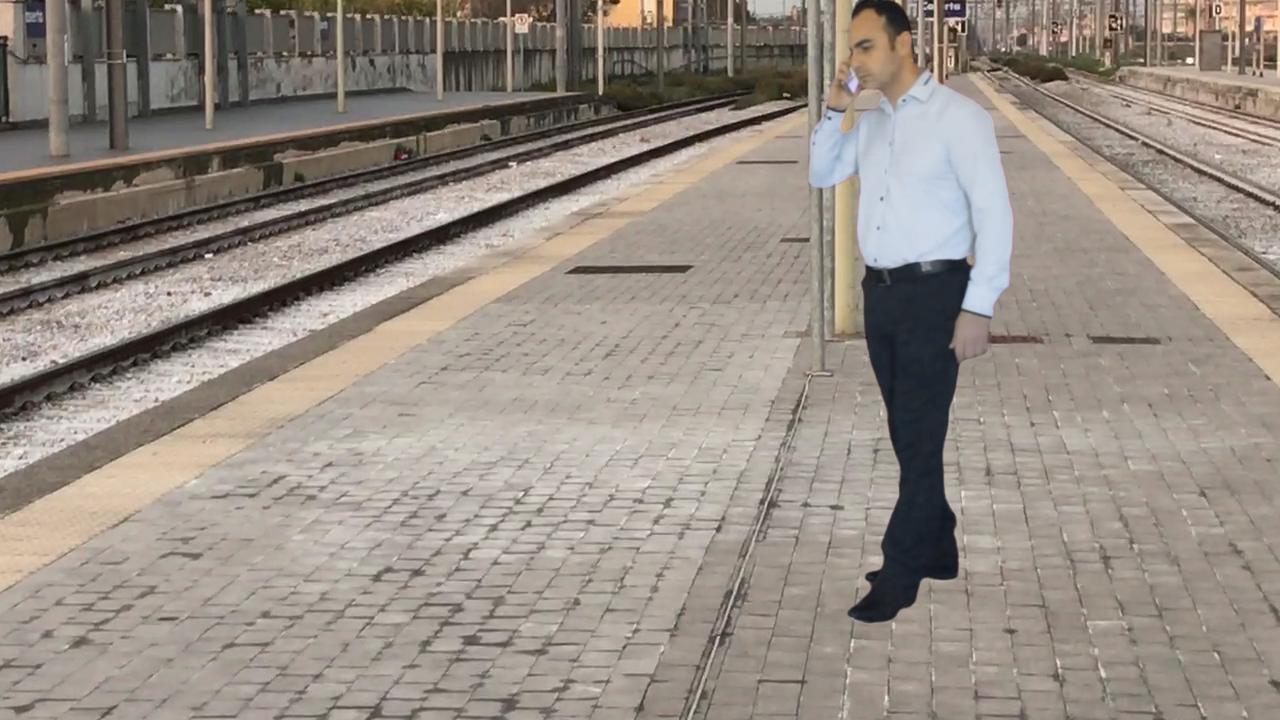} &
		\includegraphics[width=.194\textwidth]{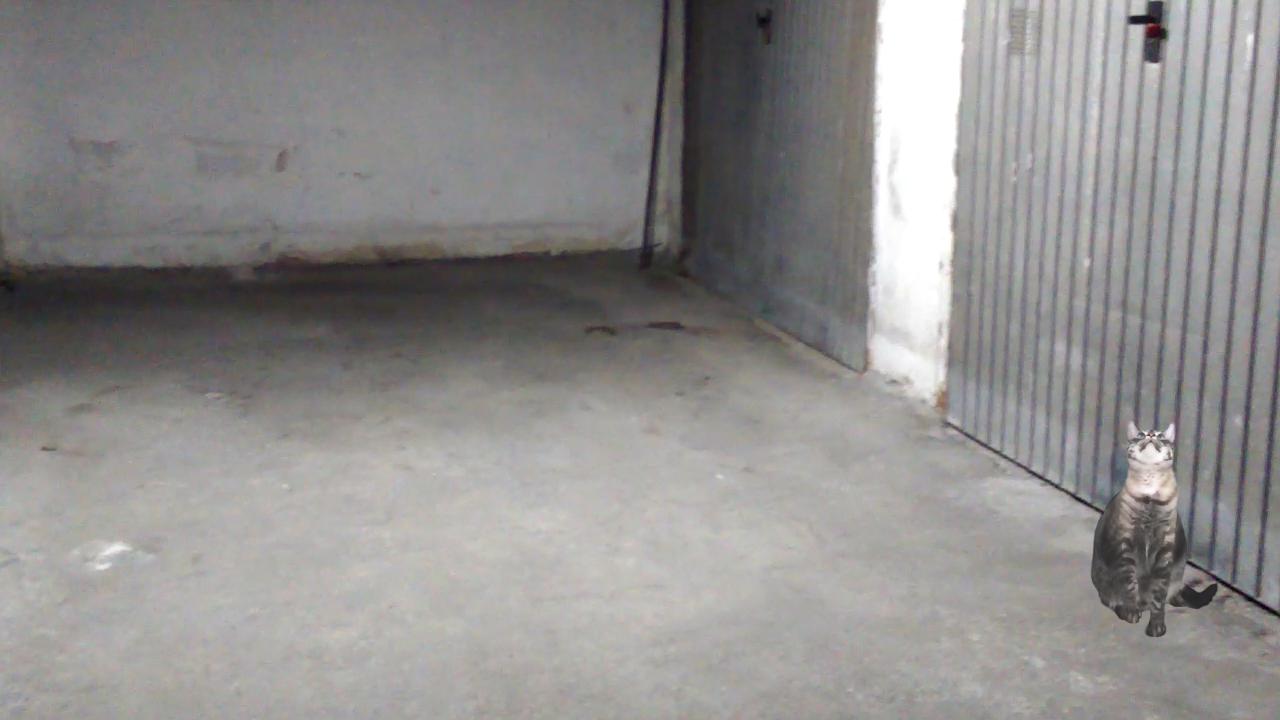} &
		\includegraphics[width=.194\textwidth]{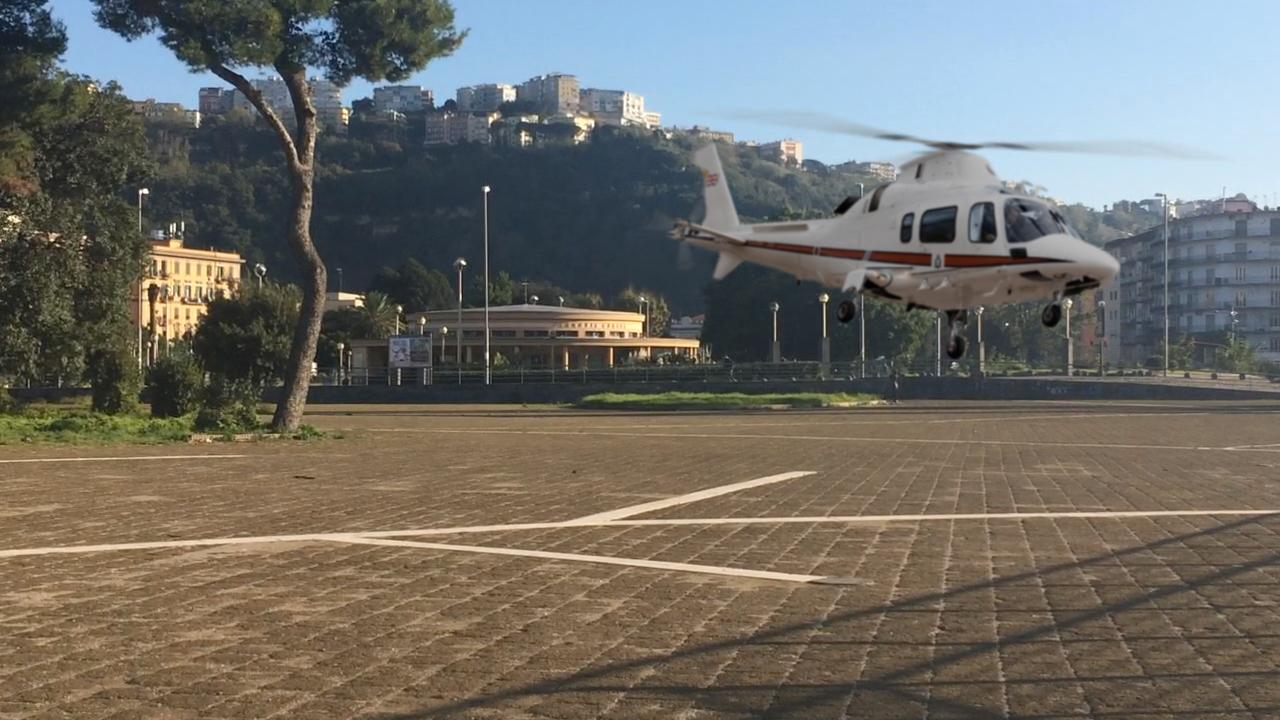} &
		\includegraphics[width=.194\textwidth]{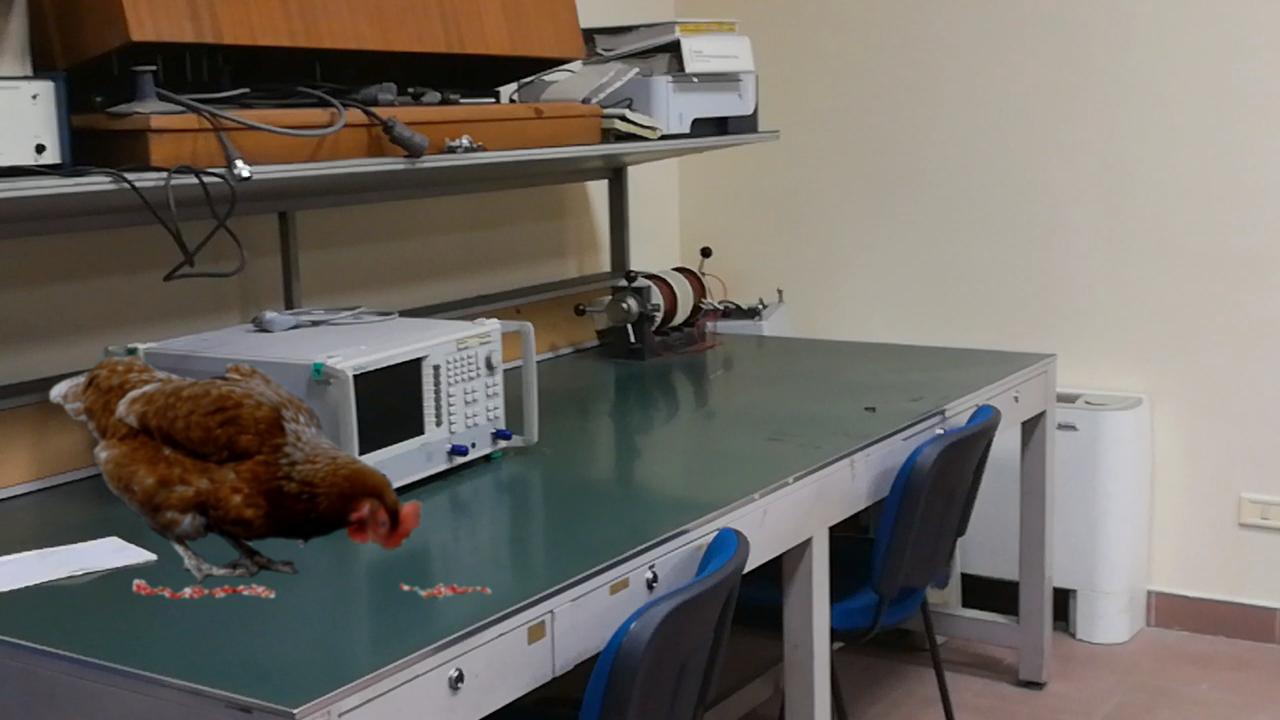} \\
		\includegraphics[width=.194\textwidth]{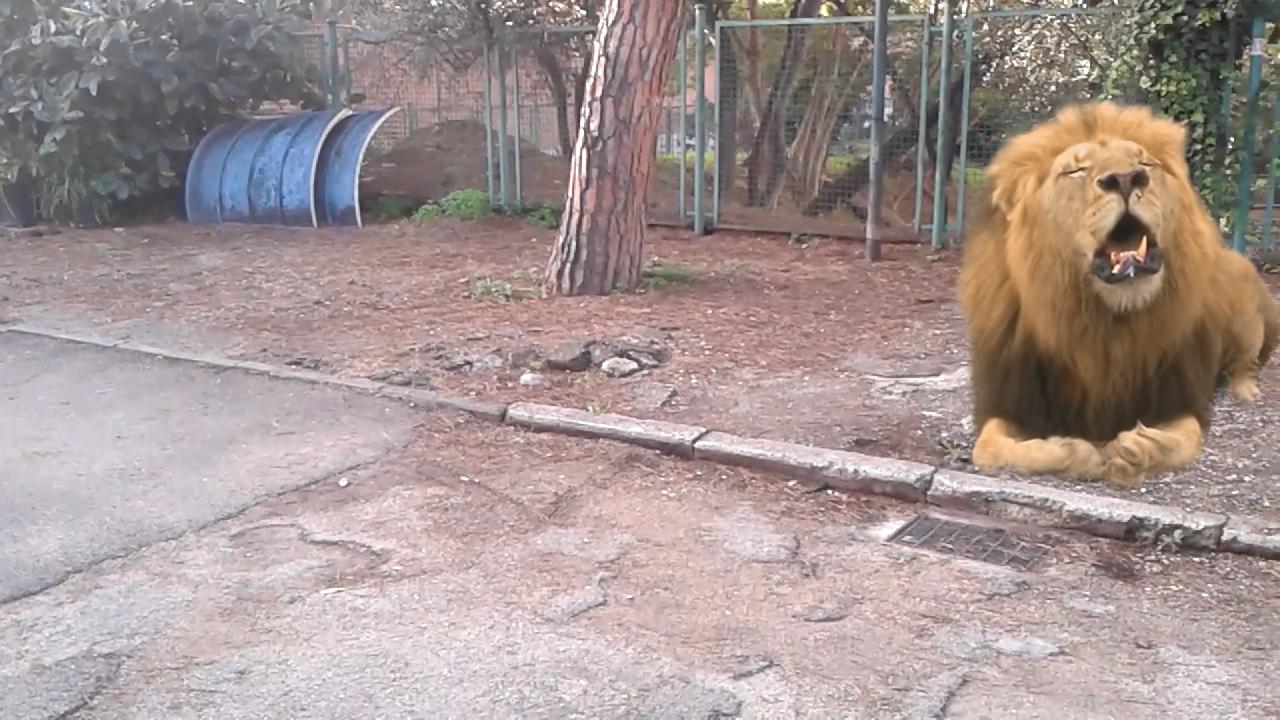} &
		\includegraphics[width=.194\textwidth]{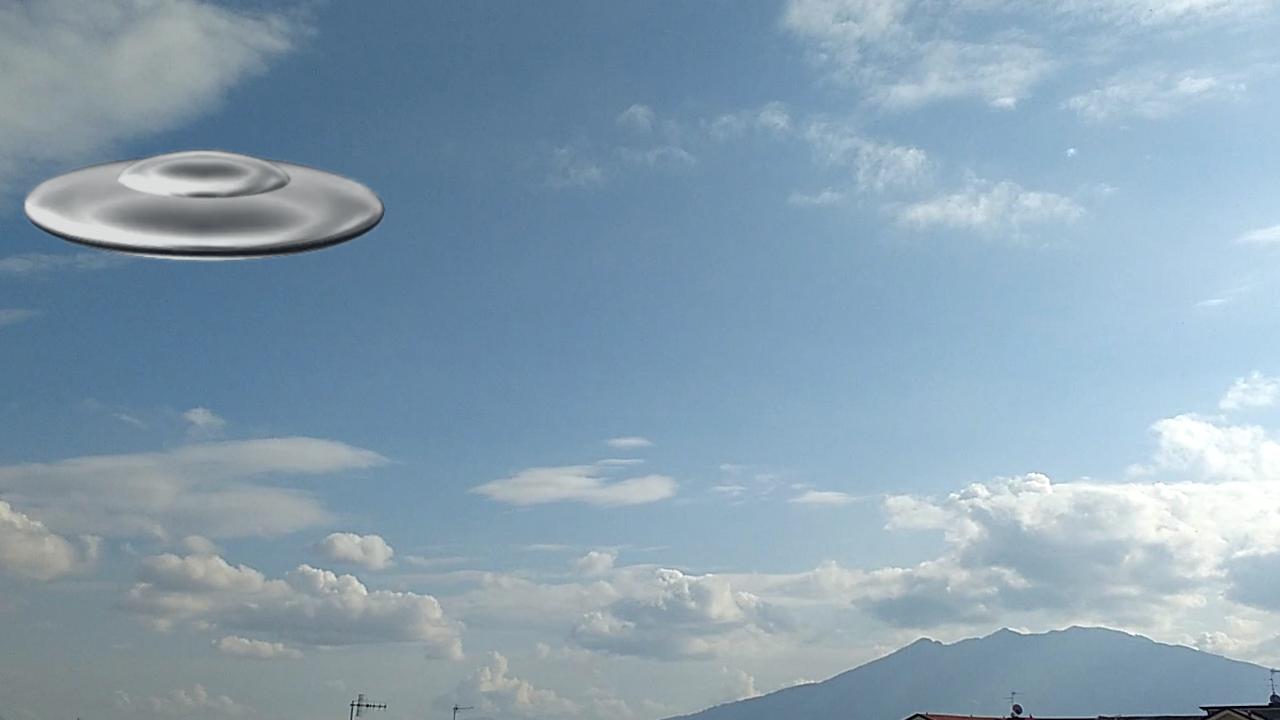} &
		\includegraphics[width=.194\textwidth]{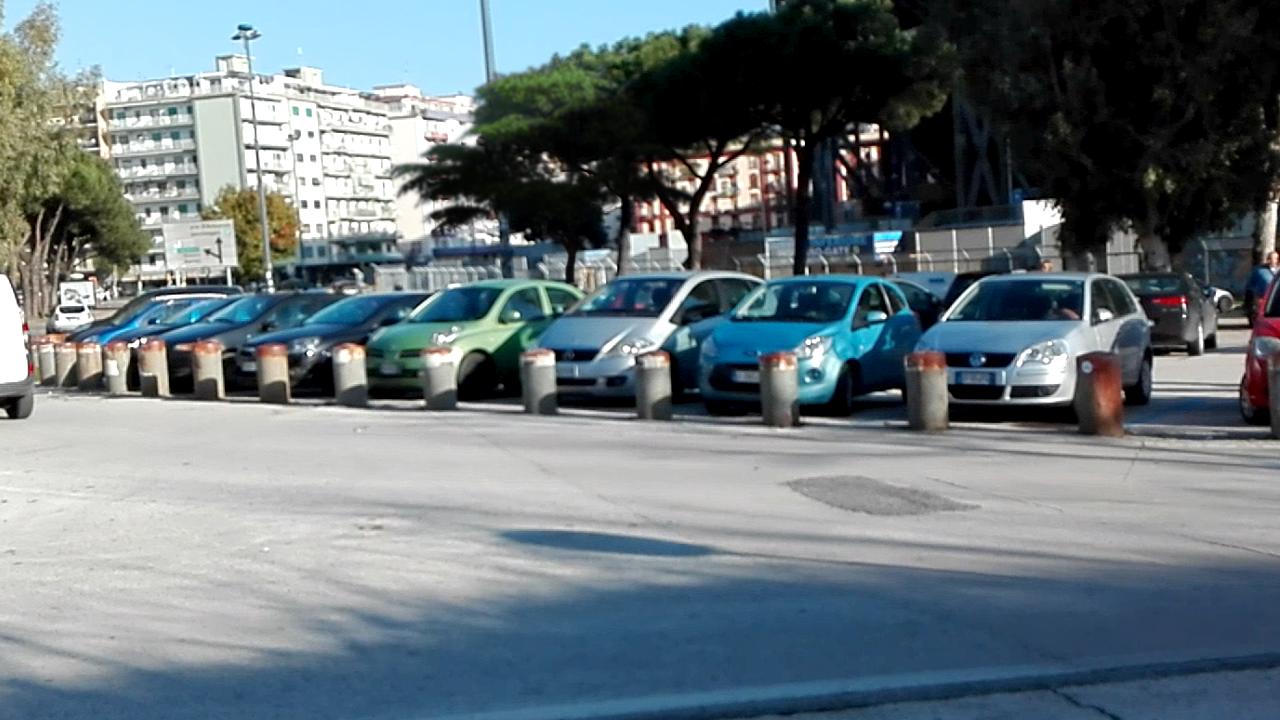} &
		\includegraphics[width=.194\textwidth]{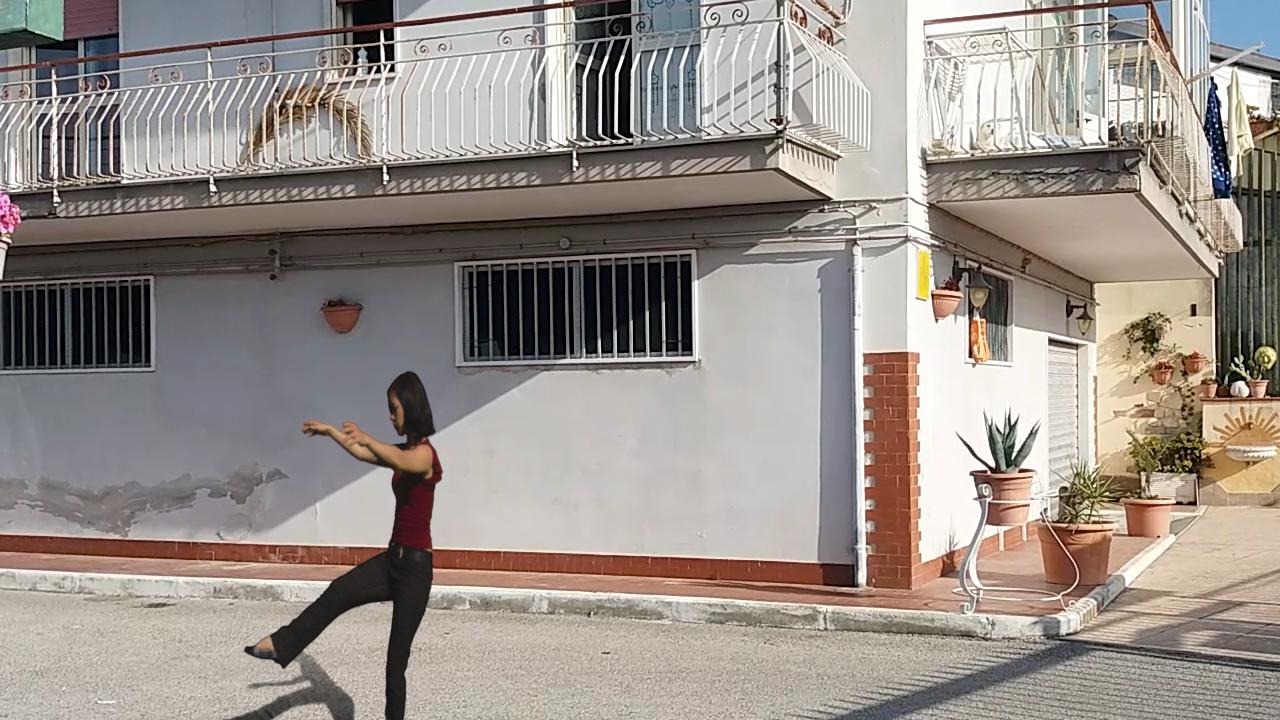} &
		\includegraphics[width=.194\textwidth]{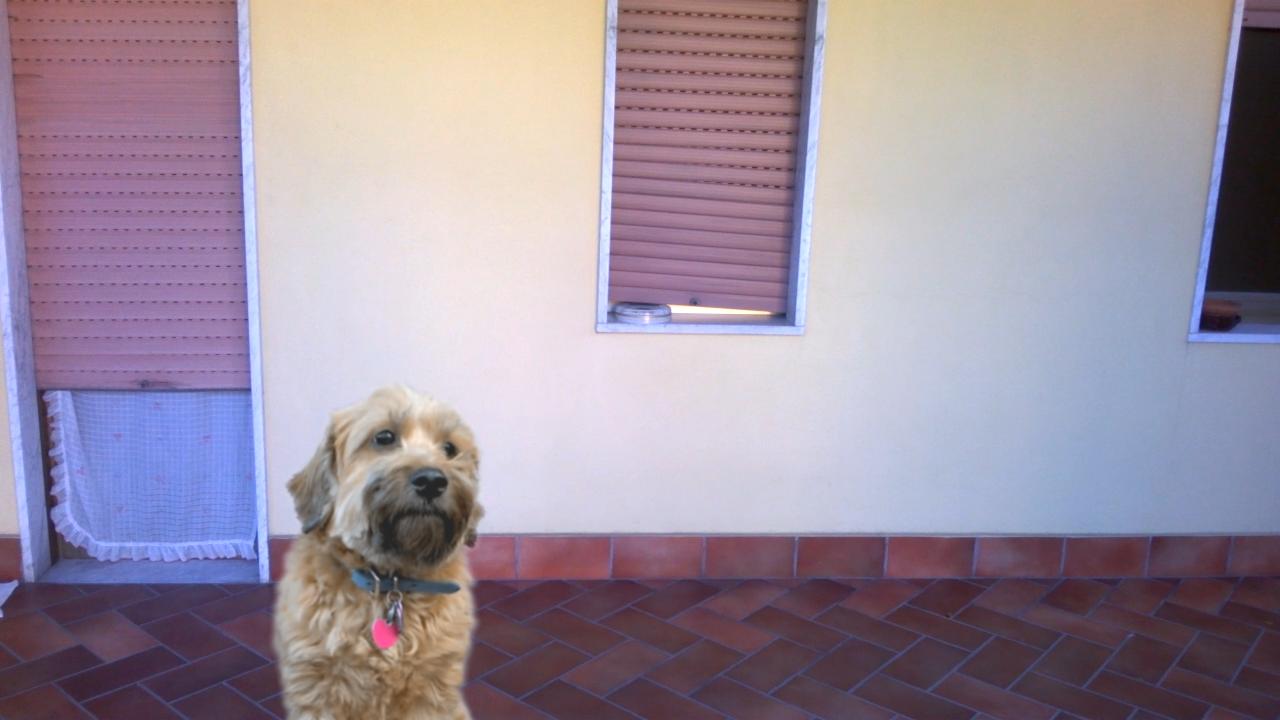} \\
	\end{tabular}
	\vspace{2mm}
	\caption{Individual frames taken from the forged videos.}
	\label{fig:dataset}
\end{figure*}

\newcommand{\ru}{\rule{0mm}{4mm}}
\begin{table}
	\footnotesize
	\centering
	\begin{tabular}{|c|c|c|c|l|} \hline
		\ru \# & name        & \# frames & \# forged & camera           \\ \hline\hline
		\ru  1 & Tank        &       335 &       191 & Nokia Lumia520   \\ \hline
		\ru  2 & Man         &       399 &       207 & Apple iPhone 7   \\ \hline
		\ru  3 & Cat         &       281 &       136 & Huawei P7 mini   \\ \hline
		\ru  4 & Helicopter  &       488 &       292 & Apple iPhone 5   \\ \hline
		\ru  5 & Hen         &       373 &       169 & Huawei P9 plus   \\ \hline
		\ru  6 & Lion        &       294 &       228 & Samsung GT I8150 \\ \hline
		\ru  7 & Ufo         &       306 &        96 & Motorola MotoG   \\ \hline
		\ru  8 & Tree        &       302 &       240 & Huawei P8lite    \\ \hline
		\ru  9 & Girl        &       371 &       162 & Samsung J5       \\ \hline
		\ru 10 & Dog         &       310 &       186 & Nokia Lumia520   \\ \hline
	\end{tabular}
	\vspace{2mm}
	\caption{Characteristics of the video dataset.}
\end{table}

The proposed method was implemented in Tensorflow using the Adam learning algorithm \cite{Kingma2015},
with learning rate 0.005, exponential decay rates for moment estimation $\beta_1=0.9$ and $\beta_2=0.999$, and $\epsilon=10^{-8}$ for numerical stability.
As already said, training was carried out on 50 frames, known in advance to be pristine.
Performance was measured instead on 100 frames.
In particular,
given the available ground truth, we computed the pixel-level true positive rates (TPR) and false positive rates (FPR)
by thresholding the heat maps (at the same level for all videos) and averaging results over all 10 videos.
Then, by varying the threshold, we obtained the receiver operating curves (ROC) shown in the following.

First of all,
we investigated the impact on performance of the number of neurons, $H$, in the hidden layer of the autoencoder.
Results are summarized in the ROCs of Fig.9 for the case of uncompressed forged videos (left) and videos downloaded from YouTube.
In both cases, the performance improves significantly by increasing $H$ from 15 to 100,
while beyond that point there is no further improvement, and even a slight impairment of performance for youtube videos.
Therefore, we set $H=100$ once and for all.
From these ROCs, the proposed method appears to perform quite well.
Performance impairs somewhat when using compressed videos, but not dramatically so.

\begin{figure}
	\centering
	\includegraphics[width=0.49\linewidth,trim=40 0 50 0, clip]{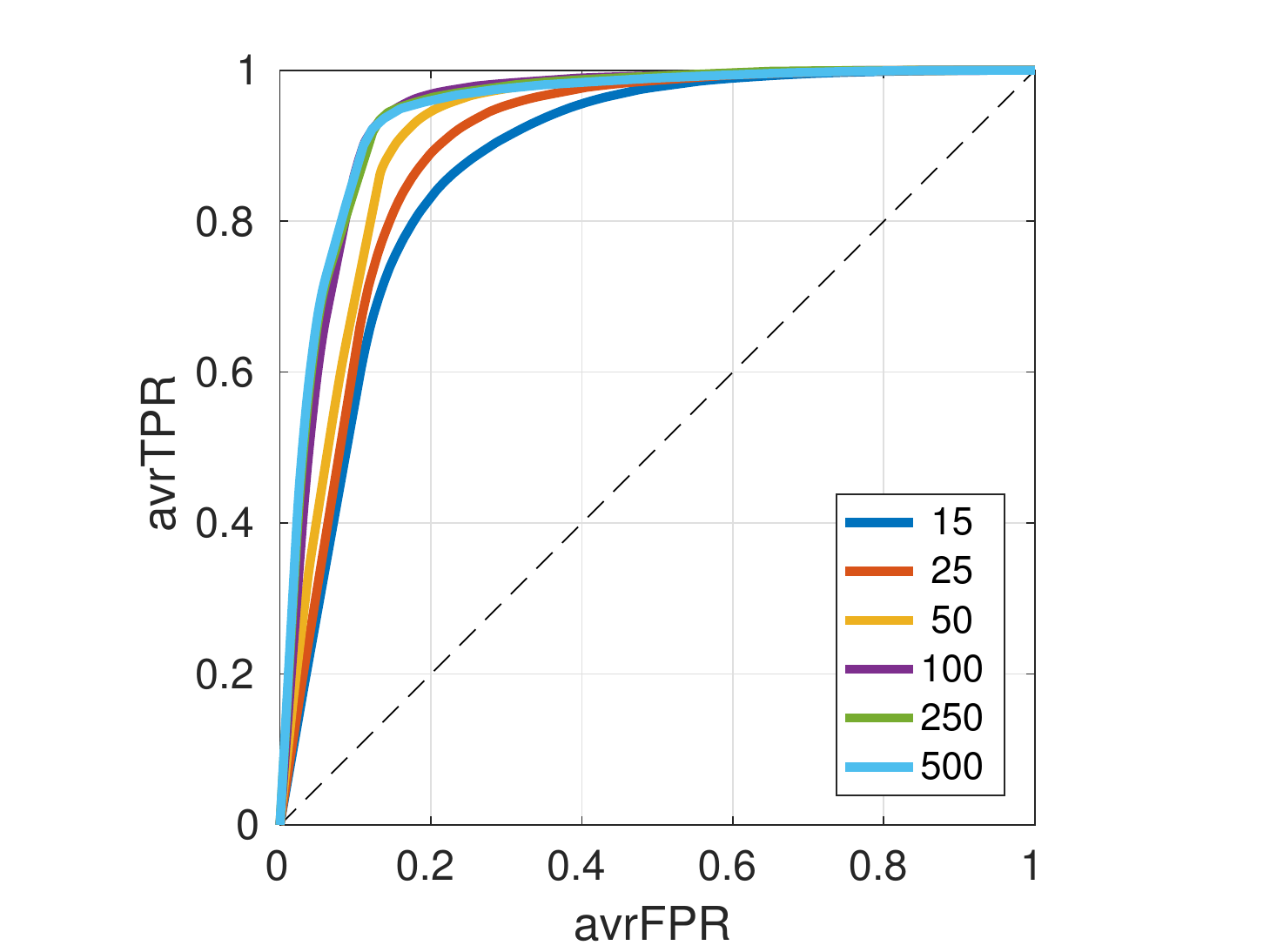}
	\includegraphics[width=0.49\linewidth,trim=40 0 50 0, clip]{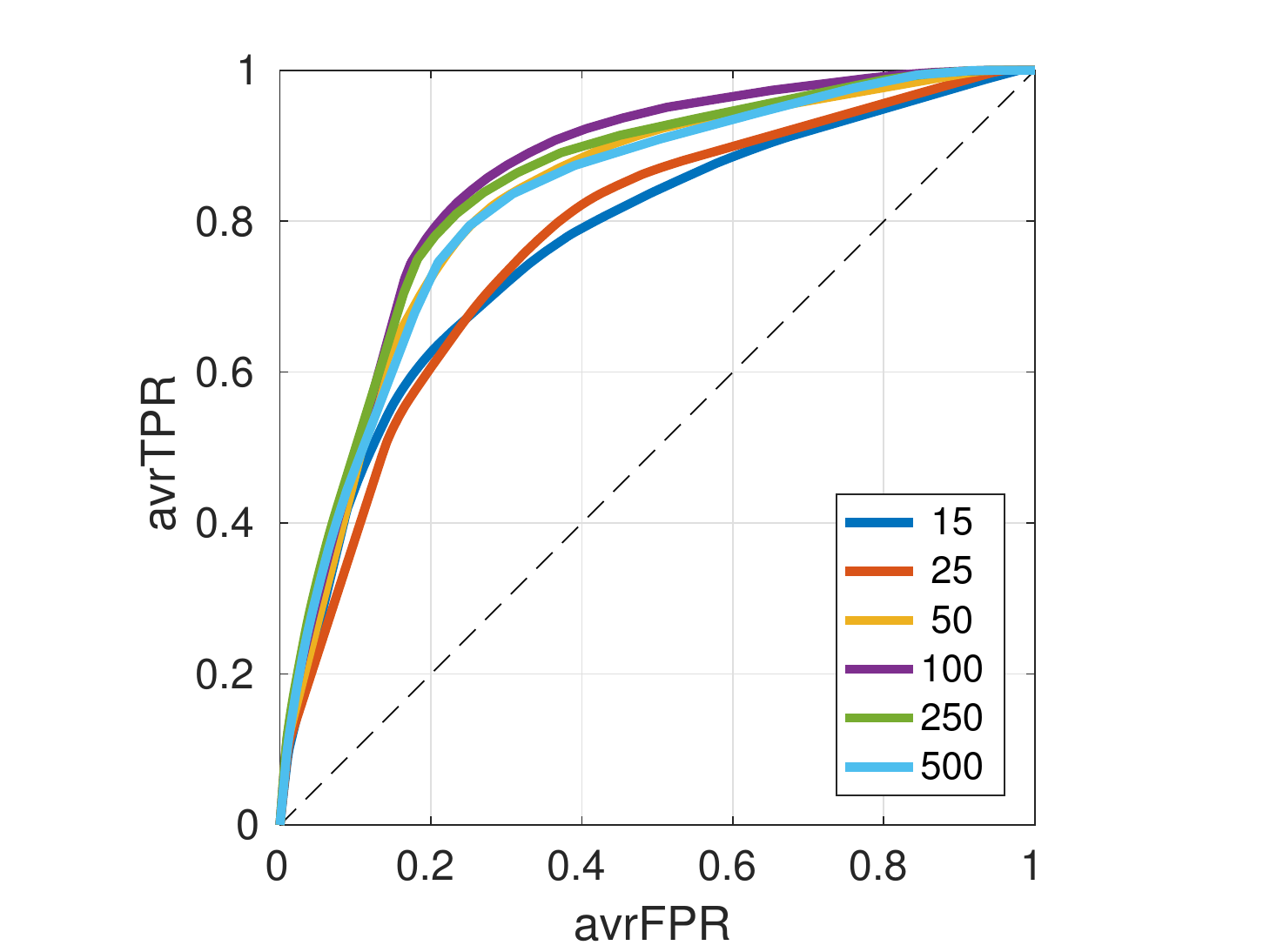}
	\caption{Performance as a function of the number of neurons in the hidden layer of the autoencoder.
    Left: uncompressed (forged) videos. Right: YouTube compressed videos.}
	\label{fig:roc_units}
\end{figure}

Then, we carried out a similar experiment to select the optimal number of unrolling steps of the LSTM recurrent neural network.
In this case, however, the impact of performance turned out to be minor,
so we set this parameter to 25 and avoid showing uninformative ROCs.
Actually,
if we renounce altogether the use of recurrent units, falling back on single-frame analysis,
the performance is only mildly affected, as shown very clearly by the ROCs of Fig.10 for both the uncompressed forged videos and YouTube ones.
This was a surprising result,
since it appears that we are taking only minimal advantage of the temporal information.
While this may be justified for uncompressed forged videos,
where very good performance are obtained,
for YouTube compressed video there is obviously room for further improvements.
A possible explanation is the fact that,
while we are using temporal information through the RNN, we are not {\em tracking} the objects through time,
and all videos are affected by significant motion.

\begin{figure}
	\centering
	\includegraphics[width=0.49\linewidth,trim=40 0 50 0, clip]{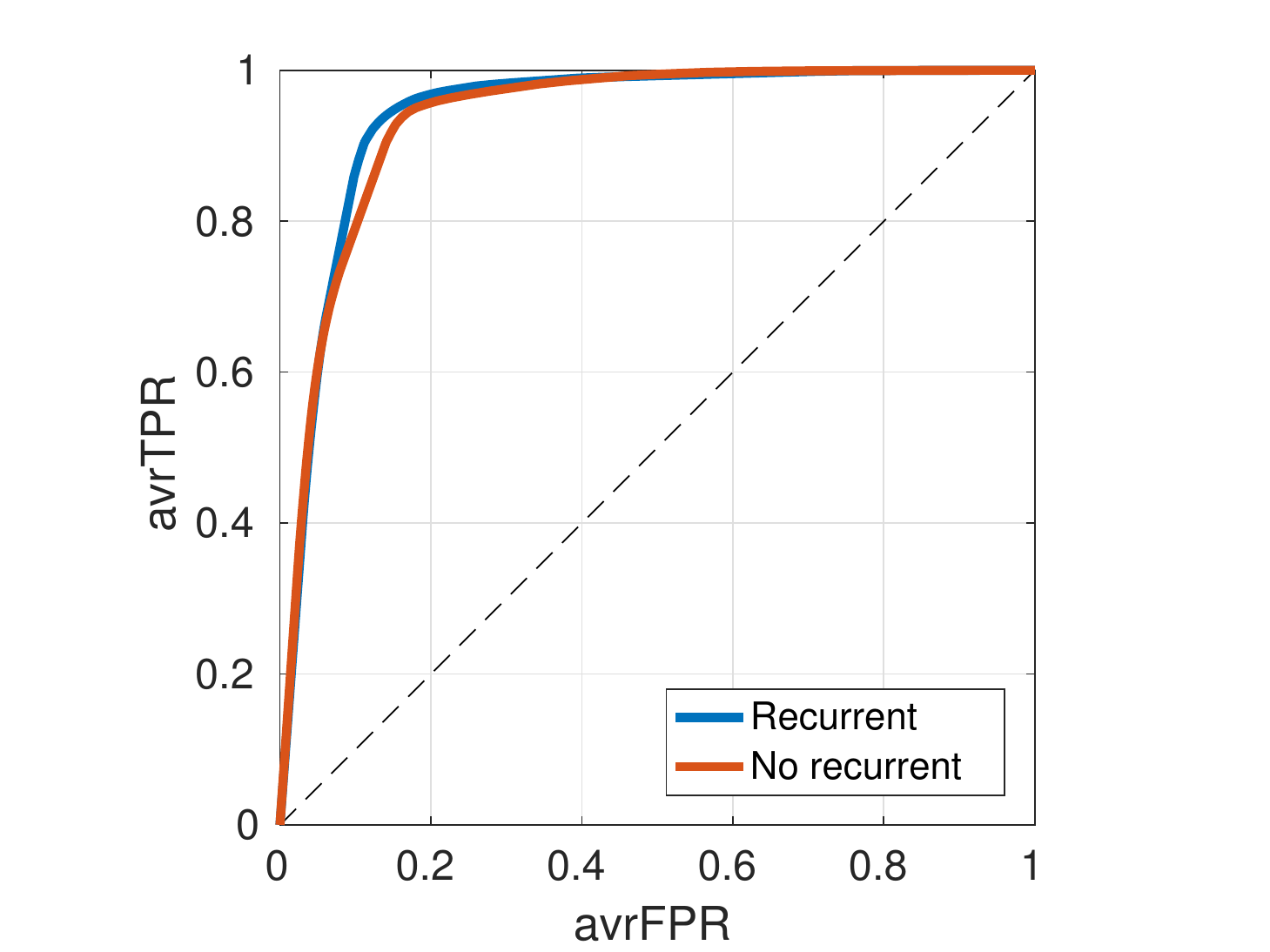}
	\includegraphics[width=0.49\linewidth,trim=40 0 50 0, clip]{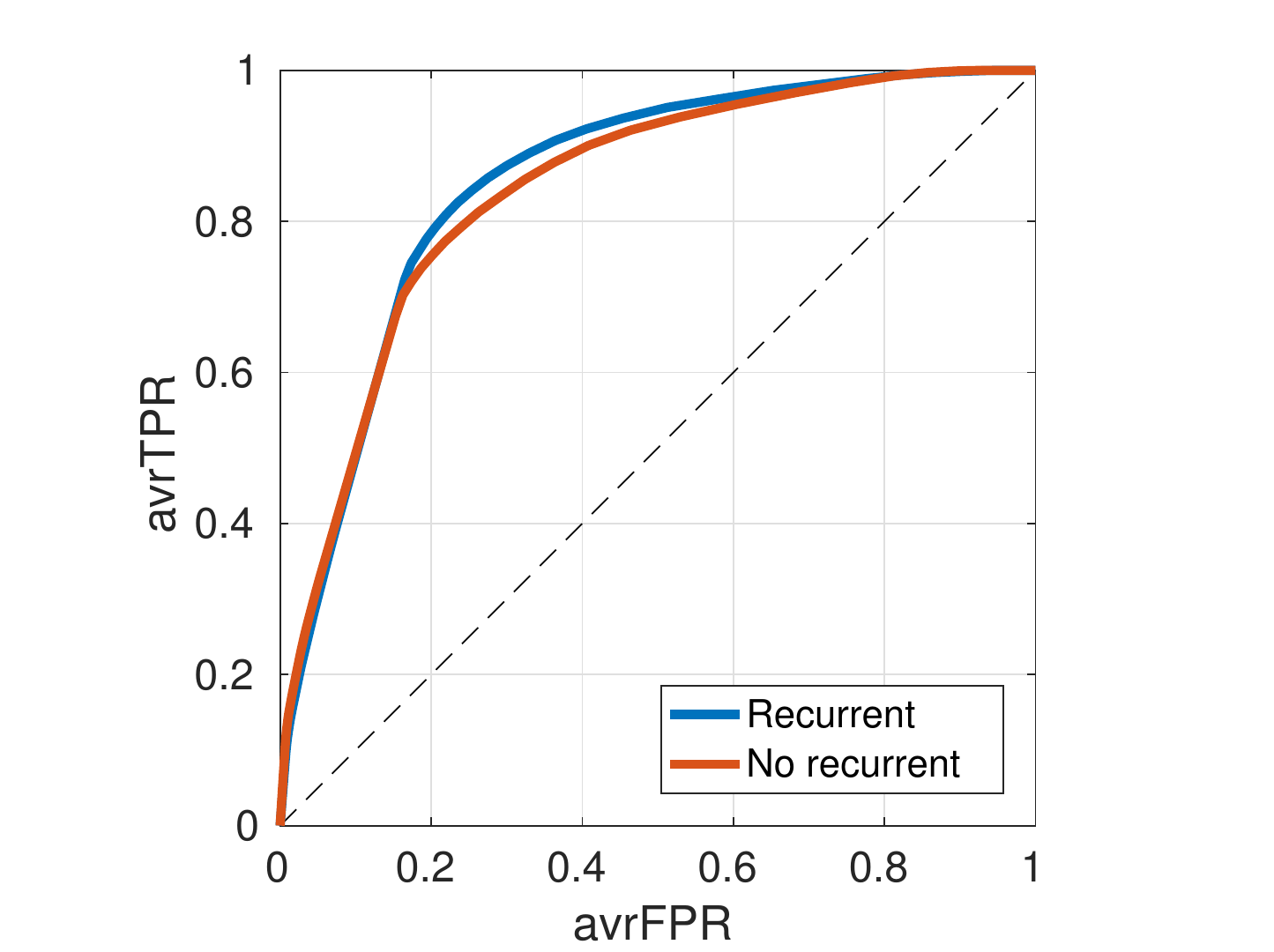}
	\caption{Performance of the proposed method with and without the use of recurrent units.
    Left: uncompressed (forged) videos. Right: YouTube compressed videos.}
	\label{fig:roc_rnn_nornn}
\end{figure}

We conclude this analysis by comparing results with some reference methods.
Unfortunately, a very limited number of candidate references can be considered and no implementation is made available by the authors.
For example,
format-based methods tailored on MPEG-2 coding, like \cite{Xu2012}, had to be discarded altogether
because modern smartphones use more advanced codecs, such as MPEG-4 or H.264.
Then, we implemented several other methods,
but in some cases could not obtain acceptable results, such as for \cite{Su2011}, proposed for chroma-key composition.
In the end, we could select as references
the model-based method proposed by ourselves in \cite{Verdoliva2014} trained on 50 frames,
called WIFS2014 from now on,
and two variants of PRNU-based detectors.
In a first case, called in-video, 50 frames of the test video are used to estimate the PRNU, as done in \cite{Mondaini2007}.
However, we also considered an ideal case, called out-video,
in which the PRNU is computed from 300 INTRA frames of uniform background so us to guarantee its reliable estimate \cite{Chuang2011}.
For both PRNU-based methods,
BM3D denoising is used \cite{Chierchia2010},
PRNU is maximum-likelihood estimated,
and detection is based on the normalized cross-correlation (NCC) computed on 128$\times$128-pixel windows,
as preliminary experiments with the more complex PCE did not show improvements.

Results are shown in Fig.11.
With uncompressed videos, the proposed method performs only slightly better than WIFS2014.
On the contrary, it provides a large improvement with respect to PRNU-based methods,
which for most videos perform only slightly better than random guessing, irrespective of the PRNU estimation quality.
Turning to the case of compressed YouTube videos,
WIFS2014 does not perform so well anymore, and the proposed method does provide a significant performance gain.
The performance of PRNU-based methods, instead, degrade further.
Interestingly, the in-video (hence, worse) PRNU estimate seems to be slightly preferable,
which raises some doubts on what these detectors are actually looking for.

\begin{figure}
	\centering
	\includegraphics[width=0.49\linewidth,trim=40 0 50 0, clip]{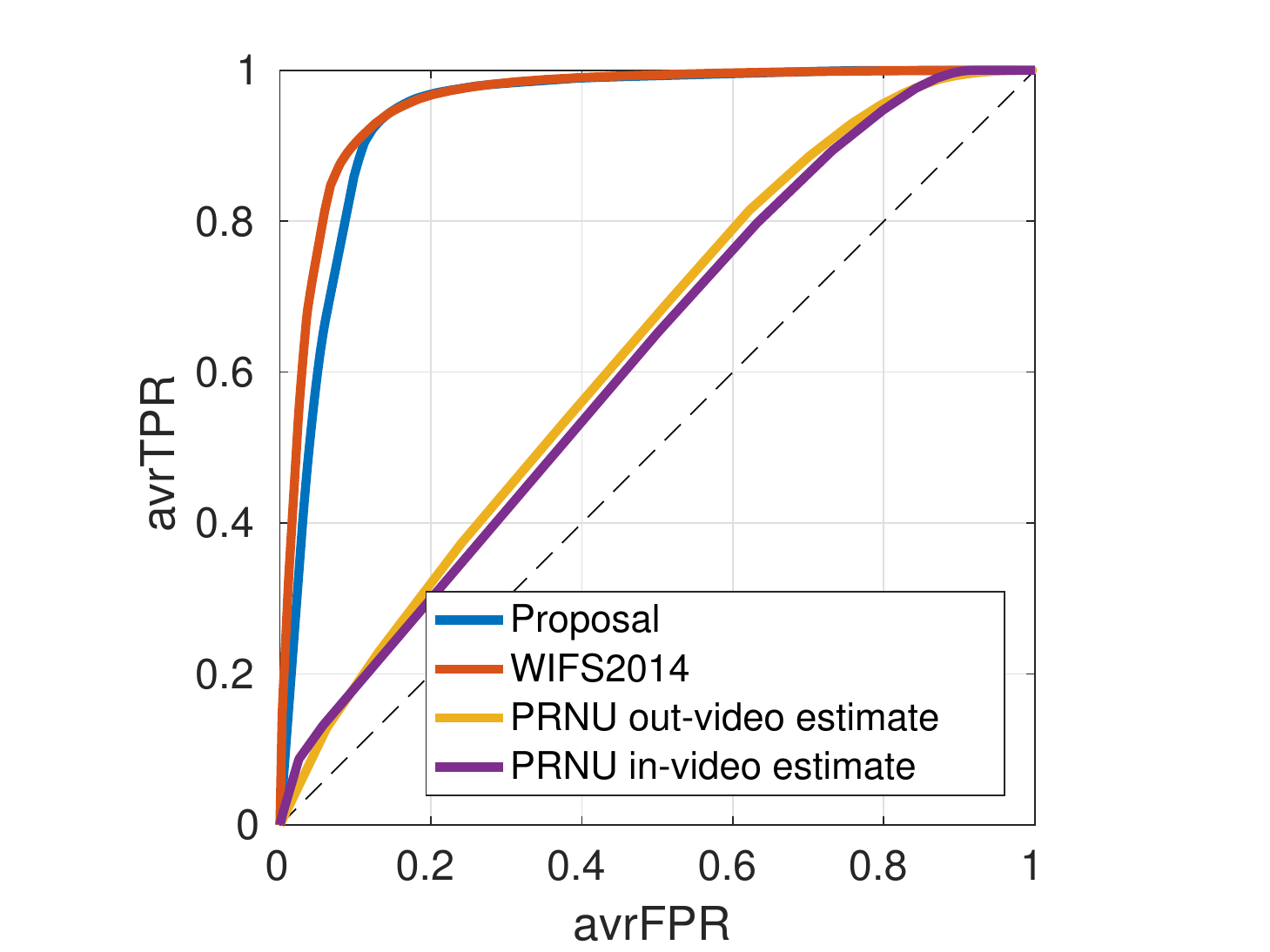}
	\includegraphics[width=0.49\linewidth,trim=40 0 50 0, clip]{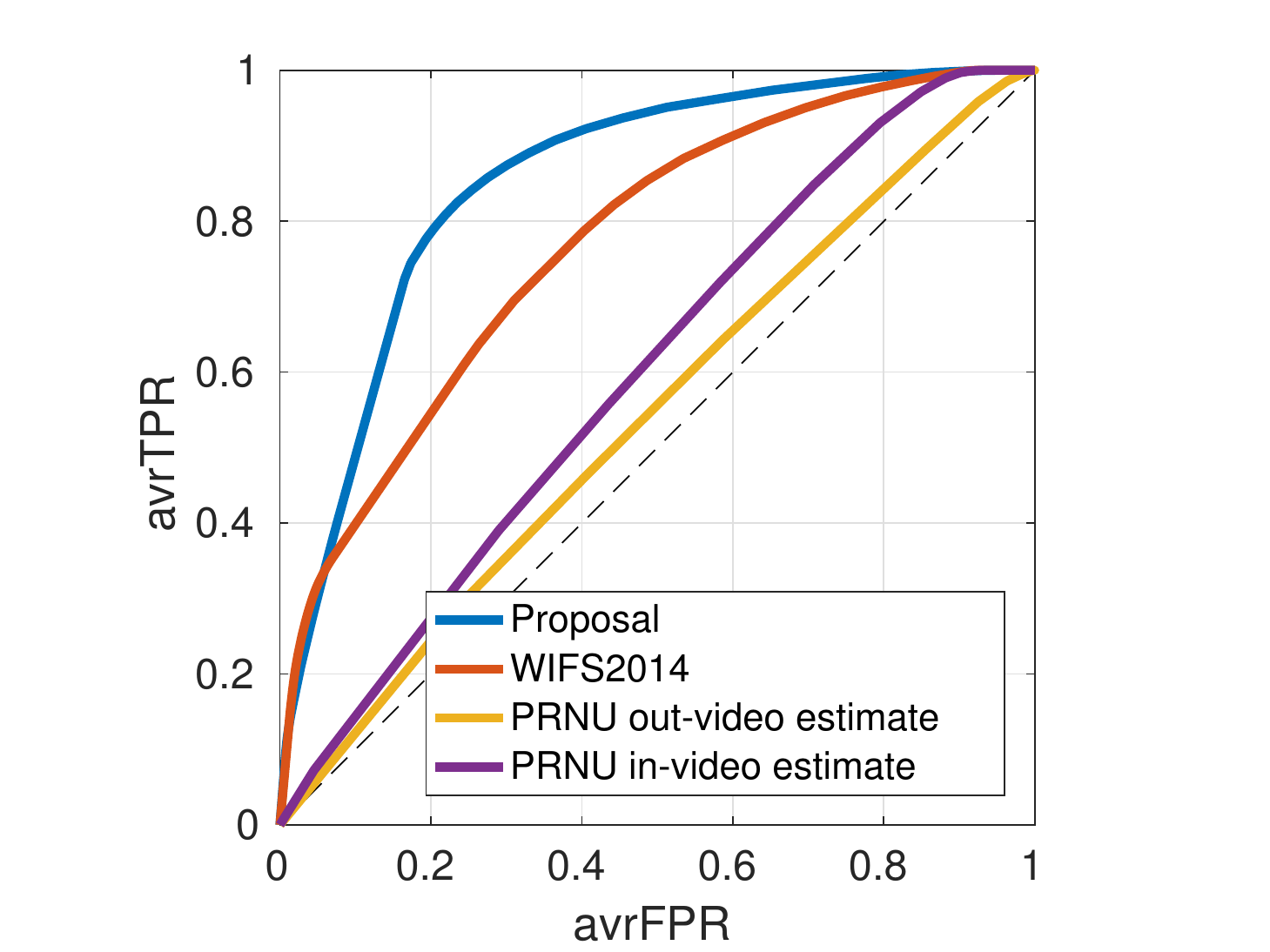}
	\caption{Comparison with state-of-the-art methods.
    Left: uncompressed (forged) videos. Right: YouTube compressed videos.}
	\label{fig:roc_comparison}
\end{figure}

Finally, we show some sample results for the Hen video, uncompressed forged (Fig.12) and YouTube compressed (Fig.13).
A short sequence of four frames is shown in the top line,
together with the heat maps output by the in-video PRNU-based method, WIFS2014, and the proposed method, in the other lines.
The visual inspection fully confirms previous observations.
WIFS2014 and the proposed method provide comparable results when no post-processing is carried out,
but the proposed method is clearly superior on the YouTube video.
In both cases, the PRNU-based heat maps look basically random.

\begin{figure*}[t]
	\centering\footnotesize
	\setlength{\tabcolsep}{1pt}
	\begin{tabular}{cccc}
		\includegraphics[width=.235\textwidth]{./figure/dataset/VID05_forged015.jpg} &
		\includegraphics[width=.235\textwidth]{./figure/dataset/VID05_forged060.jpg} &
		\includegraphics[width=.235\textwidth]{./figure/dataset/VID05_forged105.jpg} &
		\includegraphics[width=.235\textwidth]{./figure/dataset/VID05_forged150.jpg} \\
		\multicolumn{4}{c}{(a) Forged video.} \\
		\includegraphics[width=.235\textwidth]{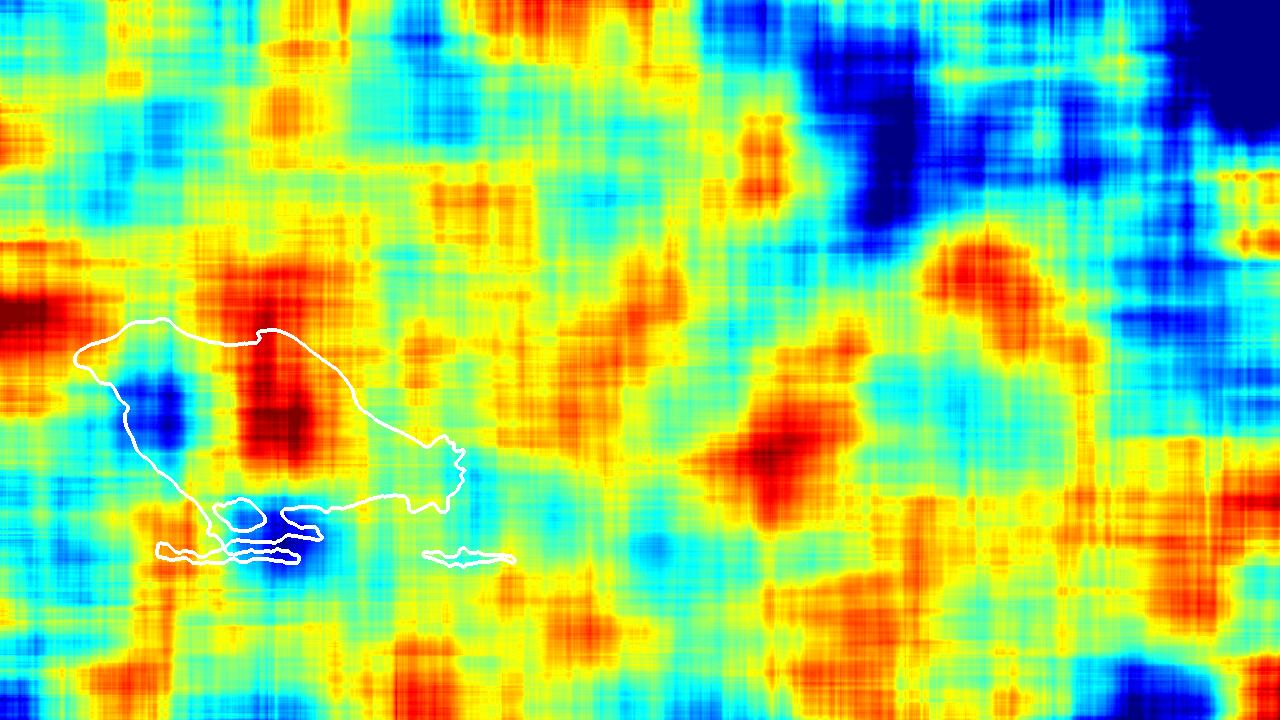} &
		\includegraphics[width=.235\textwidth]{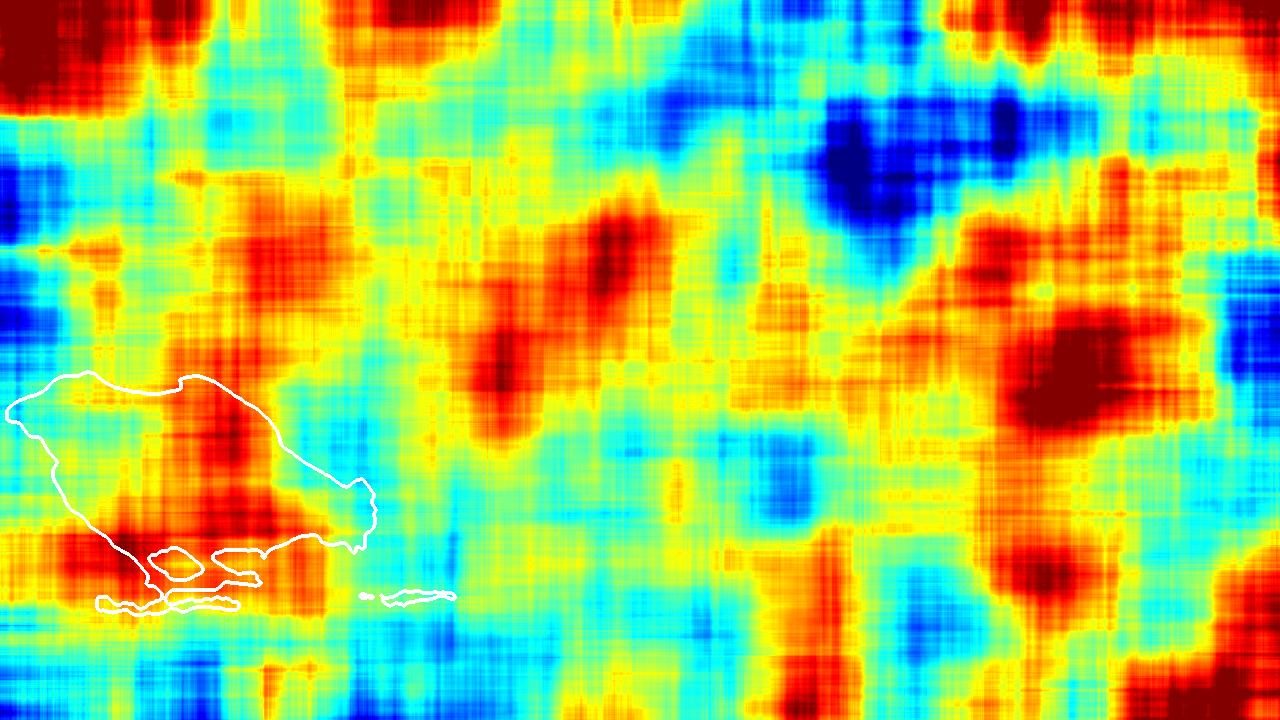} &
		\includegraphics[width=.235\textwidth]{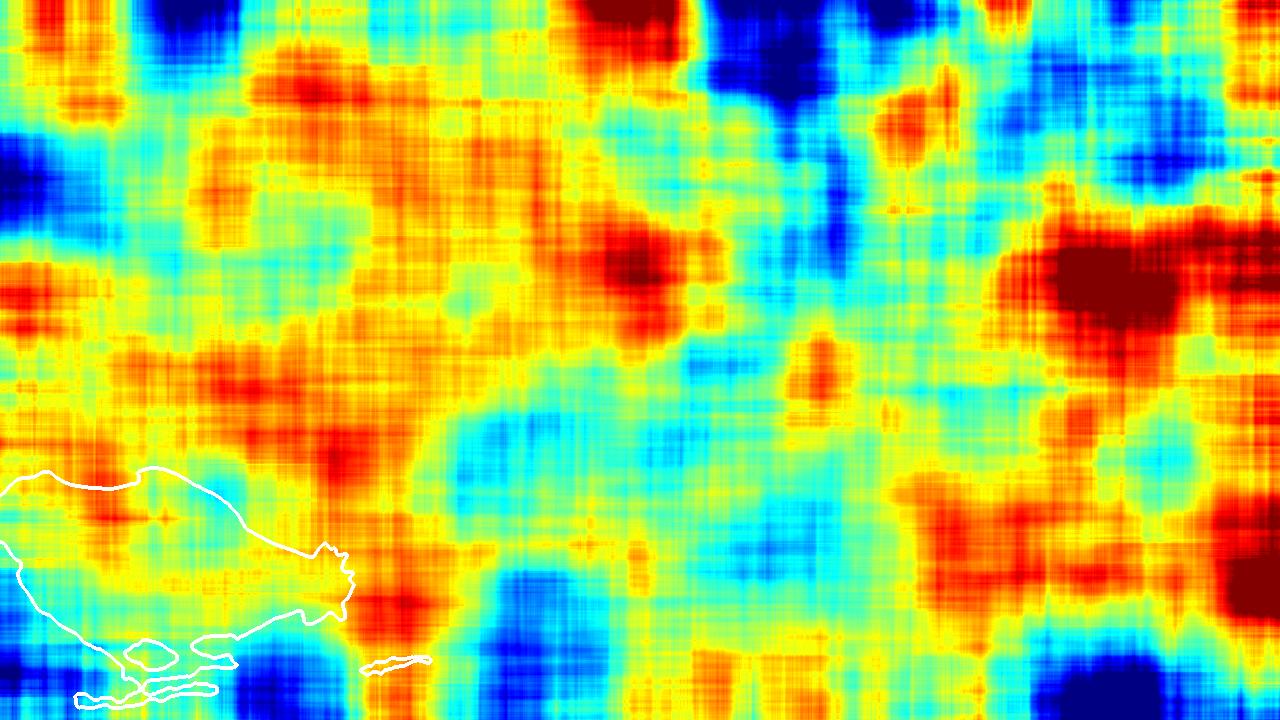} &
		\includegraphics[width=.235\textwidth]{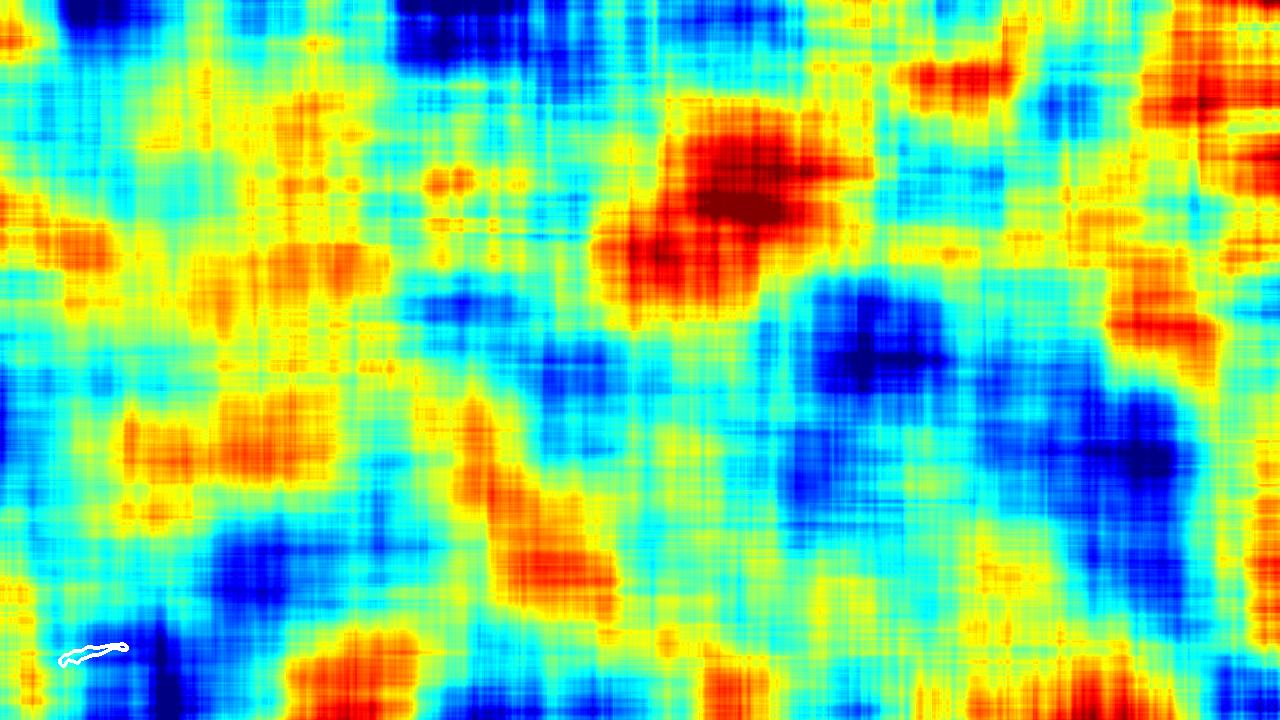} \\
		\multicolumn{4}{c}{(b) PRNU-based approach.} \\
		\includegraphics[width=.235\textwidth]{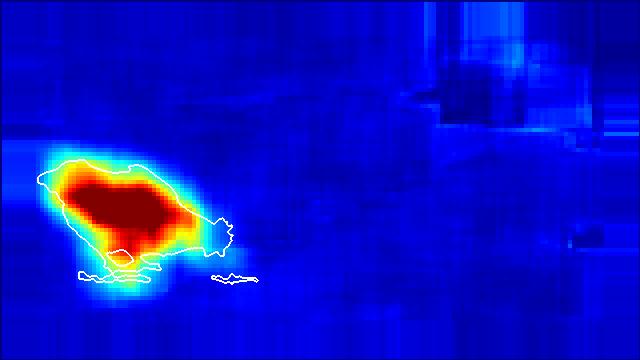} &
		\includegraphics[width=.235\textwidth]{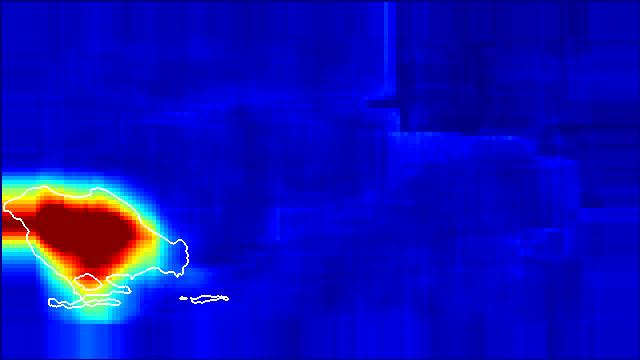} &
		\includegraphics[width=.235\textwidth]{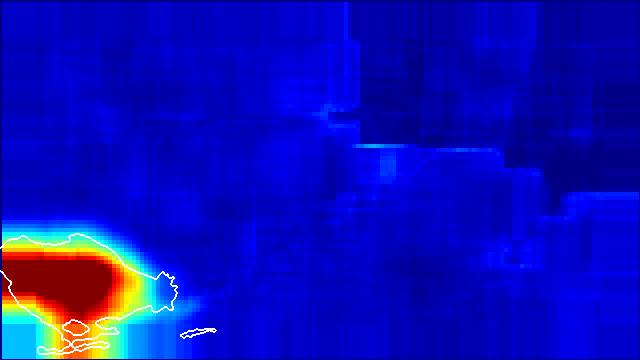} &
		\includegraphics[width=.235\textwidth]{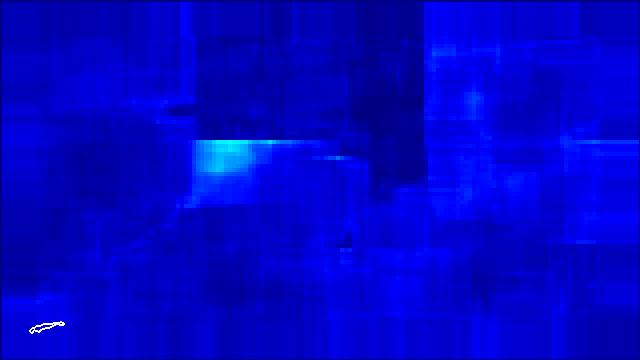} \\
		\multicolumn{4}{c}{(c) WIFS 2014.} \\
		\includegraphics[width=.235\textwidth]{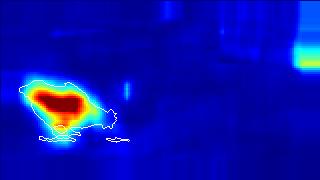} &
		\includegraphics[width=.235\textwidth]{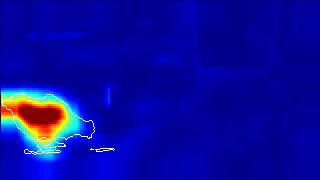} &
		\includegraphics[width=.235\textwidth]{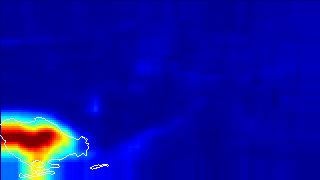} &
		\includegraphics[width=.235\textwidth]{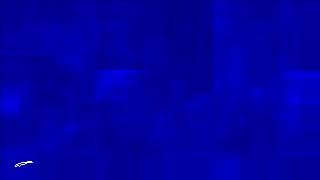} \\
		\multicolumn{4}{c}{(d) Proposed method.} \\
	\end{tabular}
	\vspace{2mm}
	\caption{Results on uncompressed (forged) video.}
	\label{fig:comp1}
\end{figure*}

\begin{figure*}[t]
	\centering\footnotesize
	\setlength{\tabcolsep}{1pt}
	\begin{tabular}{cccc}
		\includegraphics[width=.235\textwidth]{./figure/dataset/VID05_forged015.jpg} &
		\includegraphics[width=.235\textwidth]{./figure/dataset/VID05_forged060.jpg} &
		\includegraphics[width=.235\textwidth]{./figure/dataset/VID05_forged105.jpg} &
		\includegraphics[width=.235\textwidth]{./figure/dataset/VID05_forged150.jpg} \\
		\multicolumn{4}{c}{(a) Forged video.} \\
		\includegraphics[width=.235\textwidth]{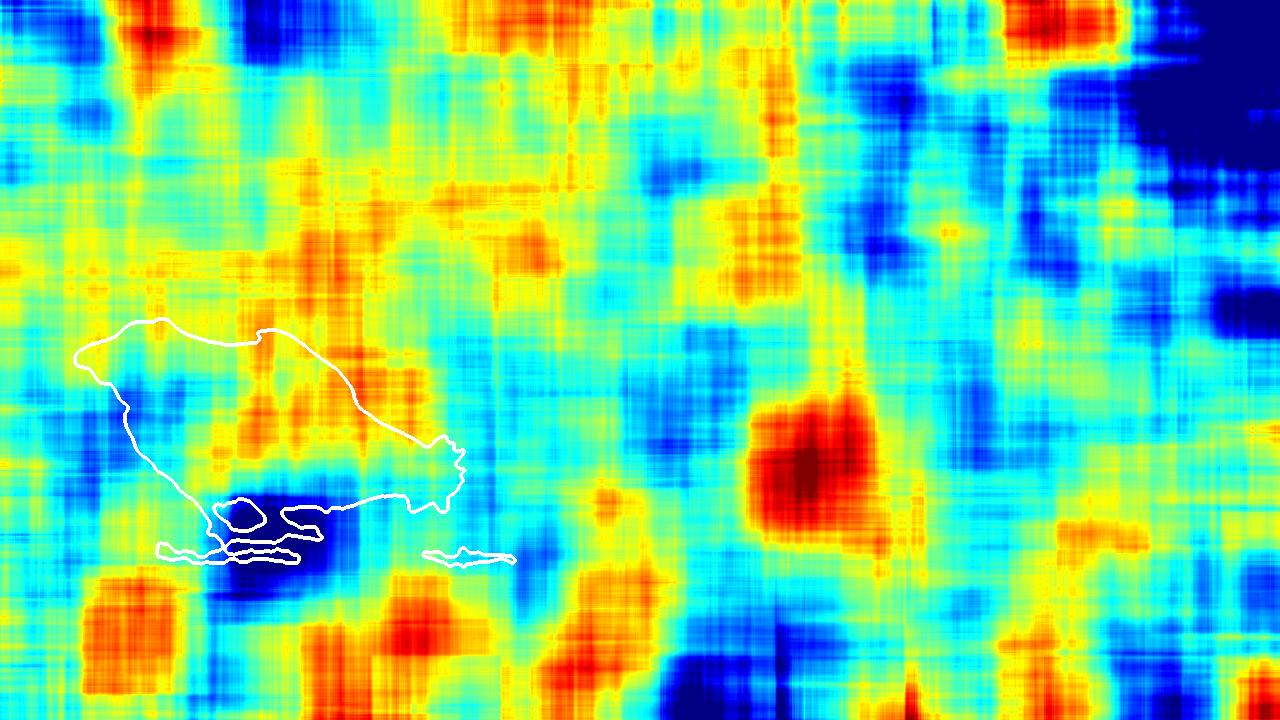} &
		\includegraphics[width=.235\textwidth]{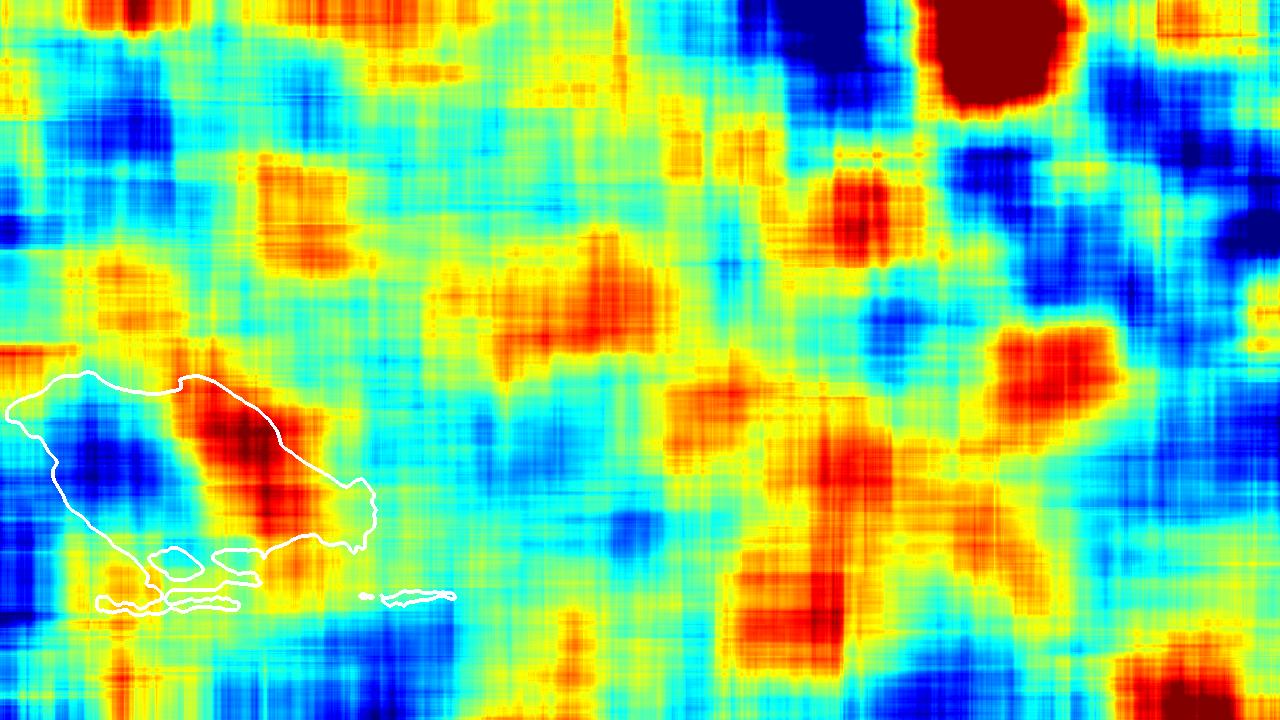} &
		\includegraphics[width=.235\textwidth]{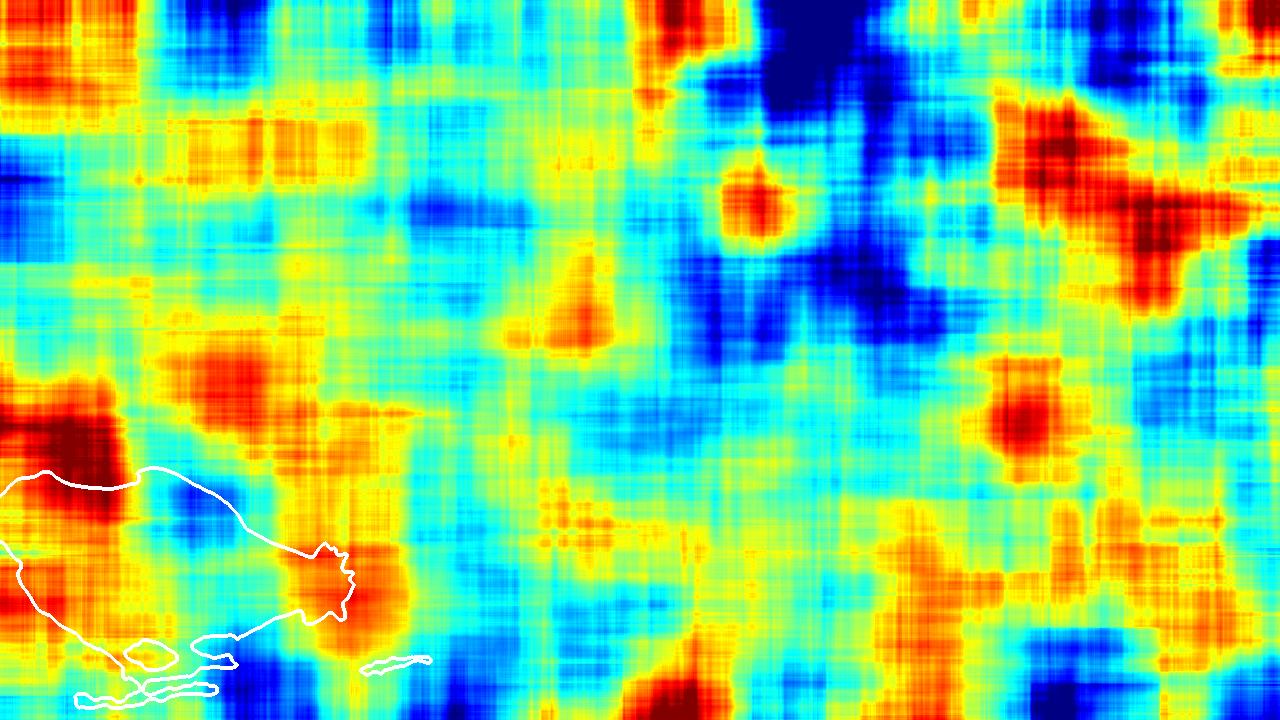} &
		\includegraphics[width=.235\textwidth]{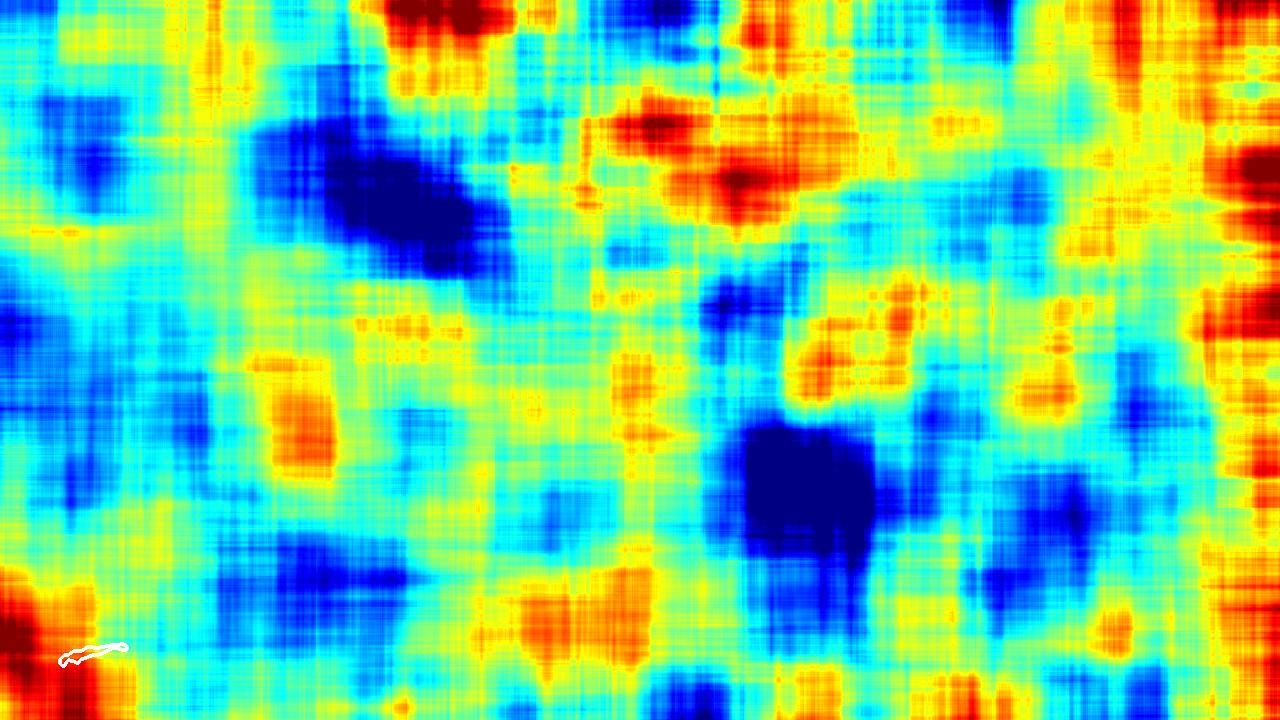} \\
		\multicolumn{4}{c}{(b) PRNU-based approach.} \\
		\includegraphics[width=.235\textwidth]{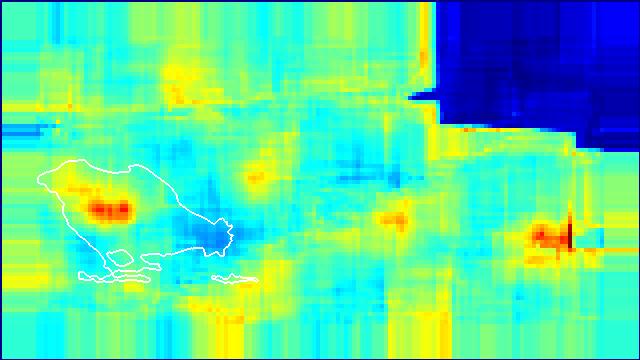} &
		\includegraphics[width=.235\textwidth]{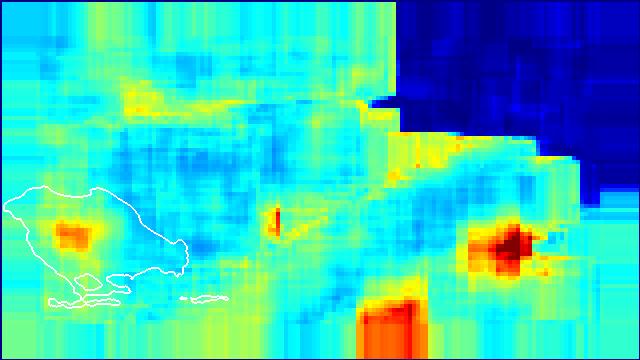} &
		\includegraphics[width=.235\textwidth]{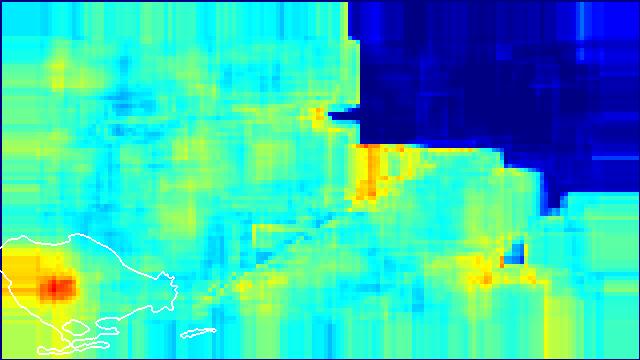} &
		\includegraphics[width=.235\textwidth]{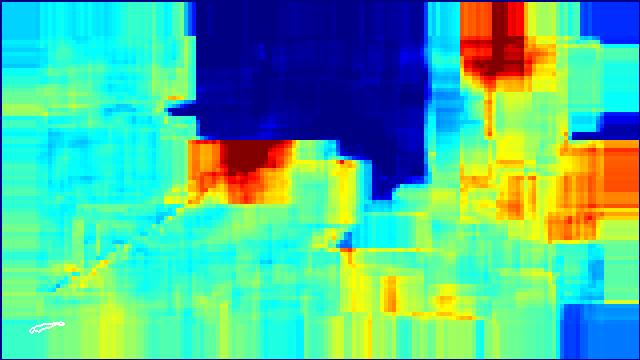} \\
		\multicolumn{4}{c}{(c) WIFS 2014.} \\
		\includegraphics[width=.235\textwidth]{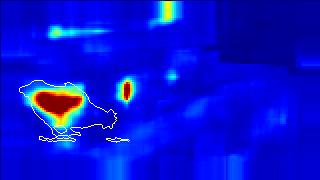} &
		\includegraphics[width=.235\textwidth]{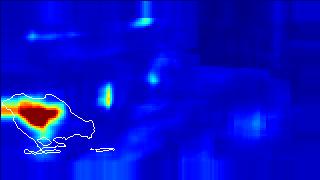} &
		\includegraphics[width=.235\textwidth]{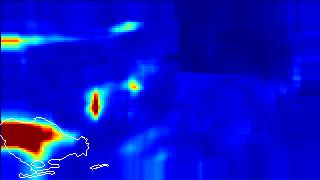} &
		\includegraphics[width=.235\textwidth]{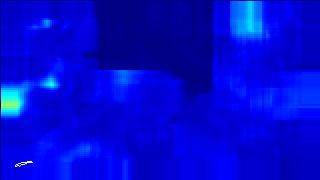} \\
		\multicolumn{4}{c}{(d) Proposed method.} \\
	\end{tabular}
	\vspace{2mm}
	\caption{Results on YouTube compressed video.}
	\label{fig:comp1c}
\end{figure*}

\section{Conclusion}

We proposed a method for video splicing detection
based on autoencoder and recurrent neural networks.
During the training phase, the autoencoder learns to reproduce
the pristine input, so that in the presence of spliced areas
the reconstruction error increases triggering detection.
Experimental results look promising,
especially on compressed videos downloaded from YouTube,
even if the use of recurrent neural networks
provides a marginal improvement.
Different directions need to be explored in the future.
A tracking algorithm should be included in order to take into account the motion
of the objects and a completely blind method,
with no need of a training phase, should be developed.
Finally, a more extensive experimental analysis should be carried out,
with different type of manipulations and post-processing operations.

\section{Acknowledgment}

This material is based on research sponsored by the Air
Force Research Laboratory and the Defense Advanced Research
Projects Agency under agreement number FA8750-16-2-0204.
The U.S. Government is authorized to reproduce and
distribute reprints for Governmental purposes notwithstanding
any copyright notation thereon. The views and conclusions
contained herein are those of the authors and should not be
interpreted as necessarily representing the official policies or
endorsements, either expressed or implied, of the Air Force
Research Laboratory and the Defense Advanced Research
Projects Agency or the U.S. Government.


\balance
\bibliography{video_refs}
\bibliographystyle{ieeetr}

%
%

\end{document}